\documentclass[sigconf, nonacm]{acmart}
\newcommand\vldbdoi{XX.XX/XXX.XX}
\newcommand\vldbpages{XXX-XXX}
\newcommand\vldbvolume{14}
\newcommand\vldbissue{1}
\newcommand\vldbyear{2023}

\newcommand\vldbtitle{\shorttitle} 
\newcommand\vldbavailabilityurl{URL_TO_YOUR_ARTIFACTS}
\newcommand\vldbpagestyle{plain} 

\usepackage{graphicx}
\usepackage{textcomp}
\usepackage{xcolor}
\def\BibTeX{{\rm B\kern-.05em{\sc i\kern-.025em b}\kern-.08em
    T\kern-.1667em\lower.7ex\hbox{E}\kern-.125emX}}
\usepackage{balance} 
\usepackage{bbm}
\usepackage{amsfonts}
\usepackage{epstopdf}
\usepackage{array}
\usepackage{booktabs}
\usepackage{color}
\usepackage{xcolor}
\usepackage{colortbl}
\usepackage{xspace}
\usepackage{multirow}
\usepackage{pifont}
\usepackage{enumitem}
\usepackage{bbding}
\usepackage{fontawesome}
\usepackage{hyperref}
\usepackage{makecell}
\usepackage{microtype}
\usepackage{caption,setspace}
\usepackage{algorithm}
\usepackage{algorithmic}
\usepackage[normalem]{ulem}
\usepackage{hyperref}

\newtheorem{lemma}{Lemma}
\newtheorem{definition}{Definition}

\begin{document}
\title{LightDiC: A Simple yet Effective Approach for Large-scale Digraph Representation Learning}







\author{
    {Xunkai Li\texorpdfstring{$^\dagger$},, 
    Meihao Liao\texorpdfstring{$^\dagger$},,
    Zhengyu Wu\texorpdfstring{$^\dagger$},,
    Daohan Su\texorpdfstring{$^\dagger$},,\texorpdfstring{\\}
    WWentao Zhang\texorpdfstring{$^{\ddagger \sharp}$},, 
    Rong-Hua Li\texorpdfstring{$^\dagger$},, 
    Guoren Wang\texorpdfstring{$^\dagger$},\texorpdfstring{\\},}
    {\texorpdfstring{$^\dagger$} BBeijing Institute of Technology, China}
    {\texorpdfstring{$^\ddagger$} MMila - Québec AI Institute, \texorpdfstring{$^\sharp$} HHEC Montréal, Canada}
    {cs.xunkai.li@gmail.com,
    mhliao@bit.edu.cn,
    Jeremywzy96@outlook.com,
    bigkdstone@foxmail.com,\texorpdfstring{\\}
    wwentao.zhang@mila.quebec,
    lironghuabit@126.com,
    wanggrbit@gmail.com}
}


\begin{abstract}
    Most existing graph neural networks (GNNs) are limited to undirected graphs, whose restricted scope of the captured relational information hinders their expressive capabilities and deployments in real-world scenarios.
    Compared with undirected graphs, directed graphs (digraphs) fit the demand for modeling more complex topological systems by capturing more intricate relationships between nodes, such as formulating transportation and financial networks.
    While some directed GNNs have been introduced, their inspiration mainly comes from deep learning architectures, which lead to redundant complexity and computation, making them inapplicable to large-scale databases.
    To address these issues, we propose LightDiC, a scalable variant of the digraph convolution based on the magnetic Laplacian.  
    Since topology-related computations are conducted solely during offline pre-processing, LightDiC achieves exceptional scalability, enabling downstream predictions to be trained separately without incurring recursive computational costs.
    Theoretical analysis shows that LightDiC utilizes directed information to achieve message passing based on the complex field, which corresponds to the proximal gradient descent process of the Dirichlet energy optimization function from the perspective of digraph signal denoising, ensuring its expressiveness.
    Experimental results demonstrate that LightDiC performs comparably well or even outperforms other SOTA methods in various downstream tasks, with fewer learnable parameters and higher training efficiency. Notably, LightDiC is the first DiGNN to provide satisfactory results in the most representative large-scale database (ogbn-papers100M).

\end{abstract}

\maketitle

\pagestyle{\vldbpagestyle}
\begingroup\small\noindent\raggedright\textbf{PVLDB Reference Format:}\\
Xunkai Li, Meihao Liao, Zhengyu Wu, Daohan Su, Wentao Zhang, Rong-Hua Li, Guoren Wang.
\vldbtitle. PVLDB, \vldbvolume(\vldbissue): \vldbpages, \vldbyear.\\
\href{https://doi.org/\vldbdoi}{doi:\vldbdoi}
\endgroup
\begingroup
\renewcommand\thefootnote{}\footnote{\noindent
This work is licensed under the Creative Commons BY-NC-ND 4.0 International License. Visit \url{https://creativecommons.org/licenses/by-nc-nd/4.0/} to view a copy of this license. For any use beyond those covered by this license, obtain permission by emailing \href{mailto:info@vldb.org}{info@vldb.org}. Copyright is held by the owner/author(s). Publication rights licensed to the VLDB Endowment. \\
\raggedright Proceedings of the VLDB Endowment, Vol. \vldbvolume, No. \vldbissue\ %
ISSN 2150-8097. \\
\href{https://doi.org/\vldbdoi}{doi:\vldbdoi} \\
}\addtocounter{footnote}{-1}\endgroup

\ifdefempty{\vldbavailabilityurl}{}{
\vspace{.3cm}
\begingroup\small\noindent\raggedright\textbf{PVLDB Artifact Availability:}\\
The source code, data, and/or other artifacts have been made available at {\url{https://github.com/xkLi-Allen/LightDiC}}.
\endgroup
}

\section{Introduction}
\label{sec: introdcution}
    Graph neural network (GNN) is a new machine learning paradig for graph-structured data, offering powerful tools for data science community to analyze and leverage information for various downstream tasks such as node-level~\cite{wu2019sgc,Hu2021ahgae,feng2022grand+}, link-level~\cite{Zhang18link_prediction1,cai2021link_prediction2,link_prediction3}, and graph-level~\cite{zhang2019graph_classification1,ma2019graph_classification2,yang2022graph_classification3}. 
    However, limitations are evident since undirected graphs fail to capture the intricate relationships between entities, leading to poor representations. 
    These issues severely restrict the performance of GNNs in analyzing complicated real-world applications and the future development of graph-based machine learning and databases.
    {For instance, when analyzing the citation network in the field of computer science (CS), the rise of AI4Industry and AI4Science in recent years has diversified and enriched citation relationships (i.e., the citation relationship occurs not only in the same field).
    Therefore, if we solely represent such data using an undirected graph, the locally directed information (e.g., CS $\to$ Biomedical or Physics $\to$ CS) is overlooked, leading to potential misguidance and erroneous model predictions.}

    To address these issues, the directed graph (digraph) is considered a promising approach for capturing advanced complexities in real-world scenarios, such as social networks~\cite{schweimer2022_directed_in_social_1,bian2020_directed_in_social_2}. 
    However, since most undirected GNNs perform poorly when being directly implemented on digraphs due to asymmetrical topology, it is a paramount necessity to design a novel directed GNN (DiGNN). 
    Recent approaches~\cite{tong2020dgcn, he2022dimpa,kollias2022nste} design two groups of learnable parameters that are separated based on the directed dichotomy to encode nodes.
    They appear to be effective intuitively, but their application is confined to small-size toy datasets and performance is unstable due to over-fitting.
    An interesting alternative solution is to define the weight-free spectral convolution by approximating the symmetric digraph Laplacian based on the original asymmetrical topology, which has been extensively studied in graph theory~\cite{chung2005spectral_graph_magnetic_laplacian1,chat2019spectral_graph_magnetic_laplacian2,shubin1994spectral_graph_magnetic_laplacian3}.
    In particular, magnetic Laplacian emerges as a powerful tool for modeling the digraphs based on the complex number~\cite{fanuel2018magLaplacian1,furutani2020magLaplacian2,guo2017magLaplacian3} due to its superior performance and interpretability. 
    
    
    Despite its effectiveness, many real-world digraphs are sparse and complex. 
    Existing DiGNNs require extra information among long-distanced nodes and vast amounts of trainable weights to learn the connection patterns, which leads to deep-coupled model architectures~\cite{tong2020dgcn,tong2020digcn,zhang2021magnet,kollias2022nste,he2022dimpa}.
    They aim to expand the receptive field (RF) of a node by aggregating information from the $K$-hop neighborhoods. 
    However, as the number of layers in the model increases, the RF grows exponentially, leading to unaffordable trainable weights and memory costs on a single machine~\cite{zhang2022pasca}. 
    Although sampling-based strategies can selectively aggregate neighbors, they are imperfect because the quality of the sampling greatly influences model performance. 
    This limits the scalability of DiGNNs, even in distributed environments, due to high communication costs.
    Moreover, undirected sampling strategies cannot be directly applied in DiGNNs due to directed edges.
    Recent advancements towards undirected scalable GNNs focus on model simplification~\cite{wu2019sgc,frasca2020sign,gamlp}, separating the feature propagation and model training to substantially reduce the computational cost.
    Meanwhile, since the undirected graph Laplacian is a special case of digraph Laplacian, the decoupled design can be directly applied to DiGNNs. 
    
    \textbf{Our contributions.}
    (1) \textit{\underline{New Perspective}}. 
    In this paper, we commence by elucidating the inherent constraints of undirected graphs in capturing intricate relationships. 
    Following this, we underscore the pivotal role played by digraphs in addressing and advancing the comprehension of real-world data science challenges.
    Subsequently, our attention turns to the issue of scalability in existing DiGNN models.
    (2) \textit{\underline{Simple yet Effective Approach}}. 
    To address scalability issues, we propose a variant of digraph convolution called LightDiC consisting of three decoupled modules.
    Specifically, LightDiC first constructs a complex Hermitian matrix called magnetic Laplacian, which is then combined with weight-free message aggregation to perform graph propagation. The above process corresponds to the proximal gradient descent process of the Dirichlet energy optimization function. 
    Finally, LightDiC collapses the complex value-based learning process into a single linear transformation for the elegance of simplicity.
    (3) \textit{\underline{High Scalability and Predictive Performance}}. 
    Extensive experiments on 7 digraph datasets demonstrate that LightDiC performs equally well or even better than other state-of-the-art baselines on various downstream tasks in terms of training efficiency (up to 358x faster), and model size (up to 16x smaller).

\begin{table*}[t]
    \caption{Algorithm analysis of existing DiGNNs in three message-passing mechanisms.
    $n$, $m$, and $f$ are the number of nodes, edges, and feature dimensions, respectively. 
    $b$ is the batch size.
    $k$ and $K$ correspond to the $k$-order proximity of neighbors and the number of times we aggregate features. 
    $\omega$ is the time complexity of computing the approximate linear rank using Gaussian random matrices or Monte Carlo sampling.
    $L$ is the number of layers in learnable classifiers trained with features and $c$ represents the complex numbers consisting of real and imaginary parts.
    }
    \footnotesize 
    \label{tab: algorithm_analysis}
    \resizebox{\textwidth}{24mm}{
    \setlength{\tabcolsep}{1.2mm}{
    \begin{tabular}{c|c|c|c|c|c}
    \midrule[0.3pt]
    Model           & Mechanism        & Pre-processing         & Training             & Inference            & Memory                        \\ \midrule[0.3pt]
    DGCN            & Directed Spat.   & $O(m^k)$               & $O(LKmf+LKnf^2)$     & $O(LKmf+LKnf^2)$     & $O(bLKf+Kf^2)$                \\
    NSTE            & Directed Spat.   & -                      & $O(LK^kmf+LK^knf^2)$ & $O(LK^kmf+LK^knf^2)$ & $O(bLK^kf+K^kf^2)$            \\
    DIMPA           & Directed Spat.   & $O(m)$                 & $O(LKk^2mf+LKk^2nf^2)$  & $O(LKk^2mf+LKk^2nf^2)$  & $O(bLKk^2f+k+Kf^2)$           \\ \midrule[0.3pt]
    DiGCN           & Mix Spat. Spect. & $O(km)$                & $O(LKmf+LKnf^2)$     & $O(LKmf+LKnf^2)$     & $O(bLKf+Kf^2)$                   \\
    DiGCN-IB        & Mix Spat. Spect. & $O(m^k)$               & $O(LKmf+LKnf^2)$     & $O(LKmf+LKnf^2)$     & $O(bLKf+Kf^2)$                   \\
    DiGCN-Appr      & Mix Spat. Spect. & $O(m)$                 & $O(Lmf+Lnf^2)$       & $O(Lmf+Lnf^2)$       & $O(bLf+f^2)$                    \\ \midrule[0.3pt]
    MGC             & Symmetric Spect. & $O(m+\log Kcm^\omega f)$ & $O(Lnc^2f^2)$           & $O(Lnc^2f^2)$           & $O(bLf+f^2)$                        \\
    MagNet          & Symmetric Spect. & $O(m)$                 & $O(Lm^cf+Ln^cf^2)$       & $O(Lm^cf+Ln^cf^2)$       & $O(bLf+f^2)$                    \\
    LightDiC (ours) & Symmetric Spect. & $O(m+Kcmf)$             & $O(ncf^2)$            & $O(ncf^2)$            & $O(bf+f^2)$                     \\ \midrule[0.3pt]
    \end{tabular}
    }}
\end{table*}

\section{Preliminaries}
\subsection{Problem Formalization}
\label{sec: preliminaries}
    We consider a digraph $\mathcal{G}=(\mathcal{V}, \mathcal{E})$ with $|\mathcal{V}|=n$ nodes, $|\mathcal{E}|=m$ edges.
    Each node has a feature vector of size $f$, stacked up in the feature matrix $\mathbf{X}\in\mathbb{R}^{n\times f}$. 
    $\mathcal{G}$ can be described by an asymmetrical adjacency matrix $\mathbf{A}(u, v),u, v \in \mathcal{V}$. $\mathbf{A}(u, v)=1$ if $(u, v) \in \mathcal{E}$ and $\mathbf{A}(u, v)=0$ vice versa.  
    $\mathbf{D}=\operatorname{diag}\left(d_1, \cdots, d_n\right) \in$ $\mathbb{R}^{n \times n}$ denotes the degree matrix of $\mathbf{A}$, where $d_i$ is the degree of node $v_i$. 
    Typical downstream tasks in digraphs include node-level and link-level.

\textbf{Node-level classification.}
    Suppose $\mathcal{V}_l$ is the labeled set, and the goal of it is to predict the labels for nodes in the unlabeled set $\mathcal{V}_u$ with the supervision of $\mathcal{V}_l$.
    For convenience, we call it Node-C.

\textbf{Link-level prediction.}
    Three typical link prediction tasks:
    (1) Direction: predict the edge direction of pairs of vertices $u, v$ for which either $(u, v) \in \mathcal{E}$ or $(v, u) \in \mathcal{E}$;
    (2) Existence: predict if $(u, v) \in \mathcal{E}$ exists in the fixed order of pairs of vertices $(u, v)$;
    (3) Three-class link classification: classify an edge $(u, v) \in \mathcal{E},(v, u) \in \mathcal{E}$, or $(u, v),(v, u) \notin \mathcal{E}$.
    For convenience, we call it Link-C.

\subsection{Directed GNNs}
\textbf{Directed spatial message-passing.}
    In the undirected cases, where the adjacency matrix $\overline{\mathbf{A}}$ is symmetric and $\overline{\mathbf{D}}$ is the degree matrix of $\overline{\mathbf{A}}$, some undirected GNNs~\cite{hamilton2017graphsage, velivckovic2017gat, xu2018jknet,frasca2020sign,huang2020cands,gamlp} follow strict spatial symmetric message-passing mechanisms to design different learnable aggregation functions, which are utilized to establish relationships among the current node and its neighbors.
    For node $u$, the $l$-th aggregator parameterized by $\mathbf{W}^{(l)}$ is represented as:
    \begin{equation}
    \label{eq:spatial_framework}
        \begin{aligned}
        \mathbf{H}_u^{(l)} =\operatorname{Aggregate}\left( \mathbf{W}^{(l)},\mathbf{H}_u^{(l-1)}, \left\{\mathbf{H}_v^{(l-1)}, \forall v \in \mathcal{N}(u)\right\}\right),
        \end{aligned}
    \end{equation}
    where $\mathbf{H}^{(0)}=\mathbf{X}, \mathbf{H}^{(l)}$ is the node embeddings in $l$-th aggregation function.
    $\mathcal{N}(u)$ represents the one-hop neighbors of $u$.
    
    To compute node embeddings in digraphs based on the asymmetrical adjacency matrix $\mathbf{A}$, DGCN~\cite{tong2020dgcn} introduces first and second-order neighbor proximity (NP) strategies to devise aggregation functions, employing two sets of independent learnable parameters for incoming and outgoing edges.
    DIMPA~\cite{he2022dimpa} increases the RF by aggregating $K$-hop neighborhoods at each model layer and leverages directed edges to independently represent source and target nodes.
    NSTE~\cite{kollias2022nste} is inspired by the 1-WL graph isomorphism test, where the information aggregation weights are tuned based on the parameterized directed message-passing process.
    DiGCN~\cite{tong2020digcn} follows the aforementioned directed spatial message-passing rules and leverages the NP to increase RF.
    Meanwhile, it notices the inherent connections between digraph Laplacian and stationary distributions of PageRank and theoretically extends personalized PageRank to construct real symmetric digraph Laplacian.
    This method advances the research on extending undirected spectral graph convolution to digraphs, enabling symmetric spectral message passing.

\noindent\textbf{Symmetric spectral message-passing.}
    Compared to the strict spatial symmetry message-passing rules in undirected graphs, some approaches~\cite{bianchi2021arma, he2021bernnet, he2022chebnetii, pmlr2022Jacobigcn, bo2023specformer} define symmetric message-passing from the spectral analysis of the undirected graph Laplacian, which is defined as $\overline{\mathbf{L}}=\overline{\mathbf{D}}-\overline{\mathbf{A}}=\mathbf{U}\mathbf{\Lambda}\mathbf{U^\mathrm{T}}$. 
    $\overline{\mathbf{L}}$ is a symmetric, positive-semidefinite matrix, and therefore has an orthonormal basis of eigenvectors $\mathbf{U}$ associated with non-negative eigenvalues $\mathbf{\Lambda}$.
    Based on this, GCN~\cite{kipf2016gcn} leverages $\mathbf{U}$ to achieve spectral convolutions on undirected graphs via the first-order approximation of Chebyshev polynomials to learn a function related to $\mathbf{\Lambda}$, which can be formally represented as
    \begin{equation}
    \label{eq:spectral_gcn}
        \begin{aligned}
        \mathbf{X}^{(l+1)}=\delta\left(\left(\widetilde{\mathbf{D}}^{-1/2} \widetilde{\mathbf{A}} \widetilde{\mathbf{D}}^{-1/2}\right)\mathbf{X}^{(l)} \mathbf{W}^{(l)}\right),
        \end{aligned}
    \end{equation}
    where $\widetilde{\mathbf{A}}=\overline{\mathbf{A}}+\mathbf{I}$, $\widetilde{\mathbf{D}}$ is the degree matrix of $\widetilde{\mathbf{A}}$, and $\mathbf{X}^{(l)}$ is $l$-layer node embeddings while $\mathbf{X}^{(0)}=\mathbf{X}$.
    In addition, $\mathbf{W}^{(l)}$ denotes the trainable weights at layer $l$, and $\delta(\cdot)$ denotes the activation function.
    
    To overcome the asymmetry of $\mathbf{A}$ on the digraphs and implement symmetric spectral message passing, 
    MagNet~\cite{zhang2021magnet} utilizes complex numbers to model directed information, it proposes a spectral GNN for digraphs based on a complex Hermitian matrix known as the magnetic Laplacian. 
    Meanwhile, MagNet uses additional trainable parameters to combine the real and imaginary filter signals separately to achieve better predictive performance.
    MGC~\cite{zhang2021mgc} adopts a truncated variant of PageRank named Linear-Rank, which designs and builds low-pass and high-pass filters for homogeneous and heterogeneous digraphs based on the magnetic Laplacian.
    The core of the above DiGNNs following symmetric spectral message-passing mechanisms lies in identifying and defining symmetric (conjugated) relations based on directed edges. 
    Subsequently, through conducting spectral analysis on these topological relations, these methods achieve a symmetric spectral message-passing process.

    \subsection{Complexity Analysis}
    \label{sec: Complexity Analysis}
    In this section, we review recent proposed DiGNNs and analyze their theoretical time and space complexity in Table ~\ref{tab: algorithm_analysis}. 
    To begin with, we clarify that the training and inference time complexity of the DGCN with $L$ layers and $K$ aggregators can be bounded by $O(LKmf+LKnf^2)$, where $O(LKmf)$ represents the total cost of the weight-free sparse-dense matrix multiplication in $\operatorname{Aggregate}\left(\cdot\right)$ from Eq.~(\ref{eq:spatial_framework}), with DGCN utilizing GCN as the mechanism of aggregation function, and $O(LKnf^2)$ being the total cost of the feature transformation achieved by applying $K$ learnable aggregator weights.
    At first glance, $O(LKnf^2)$ may appear to be the dominant term, considering that the average degree $d$ in scale-free networks is typically much smaller than the feature dimension $f$, thus resulting in $LKnf^2 > LKndf = LKmf$.
    However, in practice, the feature transformation can be performed with significantly less cost due to the improved parallelism of dense-dense matrix multiplications.
    Consequently, $O(LKmf)$ emerges as the dominating complexity term of DGCN, and the execution of full neighbor propagation becomes the primary bottleneck for achieving scalability.
    
    NSTE~\cite{kollias2022nste} performs an additional aggregation based on the $k$-order proximity in each learnable aggregator, which is bounded by $O(LK^kmf+LK^knf^2)$. 
    DIMPA~\cite{he2022dimpa} extends the RF by considering incoming and outgoing edges independently in each aggregation step $O(LKk^2mf+LKk^2nf^2)$.
    The existing methods, such as DGCN, NSTE, and DIMPA, follow directed spatial message-passing mechanisms, which inherently rely on directed edges for aggregator design, making them challenging to handle large-scale digraphs. 
    Furthermore, their use of two sets of independent learnable weights to encode source and target nodes results in a large $K$, which further exacerbates the computational costs.
    DiGCN~\cite{tong2020digcn} has three variants, which involve directed spatial and symmetric spectral message passing.
    Hence, we call them Mix Spatial Spectral.
    Among them, DiGCN and DiGCN-IB are similar to DGCN as they both use $k$-order NP as pre-processing, but the generated real symmetric adjacency matrix is different.
    DiGCN-Appr extends approximate personalized PageRank for constructing digraph Laplacian as pre-processing with time complexity of $O(m)$, which is equivalent to the undirected symmetric adjacency matrix. 
    Then, the training rule of DiGCN remains similar to DGCN, but without $K$ times aggregation.

    For methods following the symmetric spectral message passing mechanisms, both MGC~\cite{zhang2021mgc} and our proposed LightDiC follow the decoupled paradigm, MageNet~\cite{zhang2021magnet} combines the propagation and training process into a deep coupled architecture. 
    In the pre-processing, all approaches achieve a time complexity of $O(m)$ to obtain the magnetic Laplacian, with the introduction of a $O(c)$ complexity due to the complex-valued matrix.  
    Then, MGC conducts multiple graph propagation approximately with significantly larger $K$, bounded by $O(\log Kcm^\omega f)$.
    In contrast, LightDiC performs only a finite number of graph propagation with small $K$, bounded by $(OKcmf)$.
    In the training, as the magnetic Laplacian involves real and imaginary parts, the fully square recursive computation cost of MagNet grows exponentially with the increase of the number of nodes and edges, reaching $O(Lm^cf+Ln^cf^2)$. 
    In contrast, MGC performs complex-valued forward propagation with a complexity of $(Lnc^2f^2)$, while LightDiC further decouples the complex-valued matrices and reduces the computation complexity to $O(ncf^2)$ by employing the simple linear logistic regression.

\begin{figure*}[t]   
	\centering
    \setlength{\abovecaptionskip}{0.4cm}
	\includegraphics[width=\linewidth,scale=1.00]{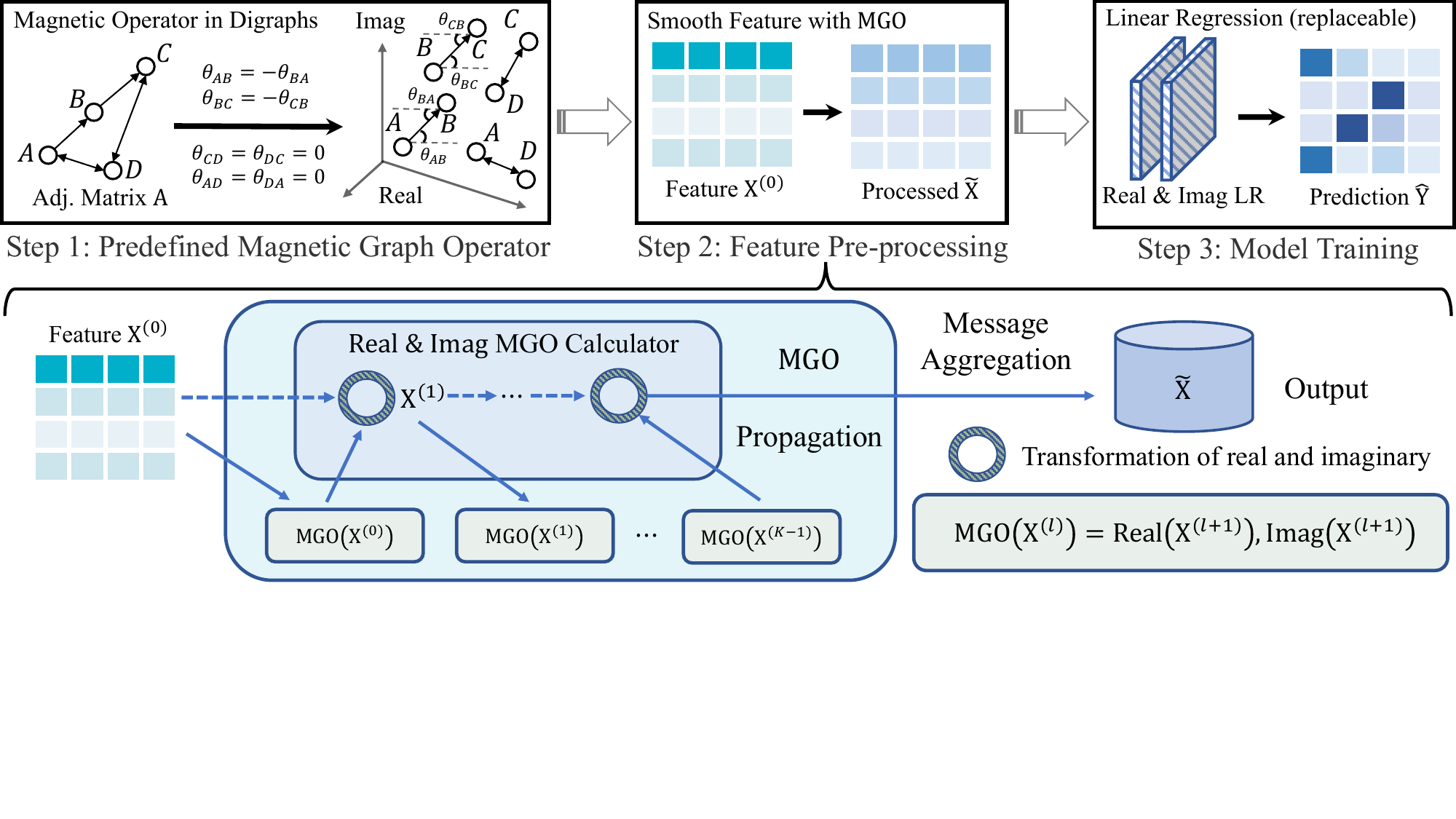}
	\caption{
     {Overview of our proposed LightDiC, including Step 1: predefined magnetic graph operator based on asymmetric digraph adjacency matrix, Step 2: feature pre-processing, and Step 3: model training with processed features.}}
	\label{fig:framework}
\end{figure*}

\section{LightDiC Framework}
     {In this section, we first introduce LightDiC, which extends digraph convolution to large-scale scenarios through three decoupled steps: 
    \textbf{Step 1}: predefined magnetic graph operator; 
    \textbf{Step 2}: feature pre-process; 
    \textbf{Step 3}: model training. 
    Remarkably, Step 1 and 2 are offline processes that are separated from the model training.
    Therefore, LightDiC performs digraph structure-related computations solely during pre-processing. 
    This allows us to train the downstream prediction process separately, avoiding the need for expensive recursive computations caused by the coupling of layer-to-layer feature propagation and transformation (details in Sec.~\ref{sec: LightDiC Pipeline} and Fig.~\ref{fig:framework}).}
    
    Subsequently, we provide essential theoretical analysis to demonstrate the applicability and interpretability of our method in real-world applications.
    Specifically, we first define the feature smoothing of digraphs from the perspective of the complex field and establish a connection with the spectral analysis of the magnetic Laplacian.
    Building upon this, we demonstrate that the feature pre-processing in LightDiC aligns with the proximal gradient descent process of the Dirichlet energy optimization function, which ensures the expressiveness of our approach (details in Sec.~\ref{sec: Theorem Analysis}).

\subsection{LightDiC Pipeline}
\label{sec: LightDiC Pipeline}
    \textbf{Predefined magnetic graph operator.}
    Since $\mathbf{A}$ is asymmetric, direct attempts to define aggregators or analyze the corresponding Laplacian typically yield high bias and complex eigenvalues.
    A preferable solution is adopting magnetic Laplacian $\mathbf{L}_m$~\cite{chung2005spectral_graph_magnetic_laplacian1,chat2019spectral_graph_magnetic_laplacian2,shubin1994spectral_graph_magnetic_laplacian3}, which is a complex-valued Hermitian matrix that encodes the asymmetric nature of a digraph via the complex part of its entries
    \begin{equation}
    \label{eq:magnetic_definition}
        \begin{aligned}
        \mathbf{L}^{(q)}_m&:=\mathbf{D}_m-\mathbf{A}_m^{(q)}=\mathbf{D}_m-\mathbf{A}_m \odot \exp \left(i \Theta^{(q)}\right),\\
        &\mathbf{A}_m(u, v):=1/2\left(\mathbf{A}(u, v)+\mathbf{A}(v, u)\right),\\
        \Theta^{(q)}&(u, v):=2 \pi q\left(\mathbf{A}(u, v)-\mathbf{A}(v, u)\right), q \geq 0,
        \end{aligned}
    \end{equation}
    where $\mathbf{D}_m$ is the degree matrix of $\mathbf{A}_m$.
    The real part in $\mathbf{L}_m(u,v)$ indicates the presence of the edge from $u$ to $v$, and the imaginary part indicates the direction.
    We follow the previous works~\cite{zhang2021magnet,zhang2021mgc,he2022msgnn} to use a $q$ parametric magnetic Laplacian to determine the strength of the directed information.
    Meanwhile, some studies~\cite{cloninger2017q_magnetic1,f2020q_magnetic2,fanuel2017q_magnetic3} clarify that different values of $q$ highlight different digraph motifs, and therefore, the appropriate value of $q$ from datasets is useful in data-driven contexts.
    Since we only consider unsigned digraphs, there exists $\cos\Theta^{(q)}\ge 0$.
    Moreover, due to the periodicity of the $\sin\Theta^{(q)}, \Theta^{(q)}\in[-\pi/2,\pi/2]$,
    we have $q\in[0,1/4]$.
    When setting $q = 0$, directed information becomes negligible.
    For $q = 1/4$, we have $\mathbf{L}_m(u, v) =-\mathbf{L}_m(v, u)$ whenever there is an edge from $u$ to $v$ only.
    Based on this, we predefine the magnetic graph operator (MGO) with self-loop ($\widetilde{\mathbf{A}}_m = \mathbf{A}_m+\mathbf{I}$) as follows
    \begin{equation}
    \label{eq:magnetic_graph_operator}
        \begin{aligned}
    \text{MGO}:=\hat{\mathbf{A}}_m=\left(\widetilde{\mathbf{D}}_m^{-1/2}\widetilde{\mathbf{A}}_m\widetilde{\mathbf{D}}_m^{-1/2} \odot \exp \left(i \Theta^{(q)}\right)\right).
        \end{aligned}
    \end{equation}
    We notice that recent studies SGC~\cite{wu2019sgc} and MagNet~\cite{zhang2021magnet} employ a similar decouple paradigm and the same MGO, respectively. 
    However, SGC solely operates on undirected graphs in the real domain and exists room for an improved message function.
    MagNet overlooks feature pre-processing and utilizes the complex-domain recursive GCN as a fundamental module during training, leaving it unsuitable for large-scale digraphs. 
    Further detailed comparisons on LightDiC, MagNet, and SGC can be found in~\cite{LightDic}.
    
    Notably, MGO is essentially a low-pass filter.
    Although a complex frequency response function can theoretically achieve better performance, we aim to propose a simple yet effective variant of digraph convolution with scalability rather than pursuing the ultimate performance with marked computational costs.
    Meanwhile, our theoretical analysis in Sec.~\ref{sec: Theorem Analysis} shows that the above MGO can still achieve both excellent performance and interpretability.
    
\noindent\textbf{Feature pre-processing.}
    Building upon the aforementioned MGO,  we can define the $K$-step weight-free feature propagation by removing the neural network $\mathbf{W}$ and nonlinear activation $\delta(\cdot)$
    \begin{equation}
    \label{eq:magnetic_feature_prop}
        \begin{aligned}
        \widetilde{\mathbf{X}}^{(K)}=\hat{\mathbf{A}}_m^{K}\widetilde{\mathbf{X}}^{(0)},\;\widetilde{\mathbf{X}}^{(K)}:=\text{Real}\left(\widetilde{\mathbf{X}}^{(K)}\right),\text{Imag}\left(\widetilde{\mathbf{X}}^{(K)}\right),
        \end{aligned}
    \end{equation}
    where $\text{Real}(\widetilde{\mathbf{X}}^{(0)})=\text{Imag}(\widetilde{\mathbf{X}}^{(0)})=\mathbf{X}$.
    Notably, the neighborhood expansion of the complex domain consists of both real part $\text{Real}(\cdot)$ and imaginary part $\text{Imag}(\cdot)$.
    After $K$-step feature propagation shown in Eq.~(\ref{eq:magnetic_feature_prop}), we correspondingly get a list of propagated features (messages) under different steps: $[\widetilde{\mathbf{X}}^{(0)},\widetilde{\mathbf{X}}^{(1)},\dots,\widetilde{\mathbf{X}}^{(K)}]$.
    Building upon this, we propose to encode multi-scale directed complex structural information in a weight-free manner $\operatorname{Message-Aggregation}(\cdot)$
        \begin{equation}
    \label{eq:magnetic_feature_aggre}
        \begin{aligned}
        \widetilde{\mathbf{X}}=\operatorname{Message-Aggregation}\left(\widetilde{\mathbf{X}}^{(0)},\widetilde{\mathbf{X}}^{(1)},\dots,\widetilde{\mathbf{X}}^{(K)}\right).
        \end{aligned}
    \end{equation}
    This decision to exclude weighted manner is driven by the need to handle the real and imaginary components as distinct entities.
    Integrating a weighted fusion of these complex value-based propagated features would impose considerable training complexity.
    Furthermore, the effective weighted aggregation of these components poses a unique challenge, hinging heavily on well-designed learning architectures and consequently leading to a marked increase in computation costs.
    This diverts from our initial aspiration to develop a simple yet efficient approach for large-scale digraph representation learning. 
    Therefore, we choose to opt for a weight-free strategy in encoding structural insights in the complex domain. 

     {From a theoretical standpoint, the small eigenvalues correspond to smoother eigenvectors within the eigendecomposition of MGO (see Sec.~\ref{sec: Theorem Analysis}).
    Therefore, we use MGO to smooth node features by Eq.~(\ref{eq:magnetic_feature_prop}), which can be regarded as the proximal gradient descent process of the Dirichlet energy.
    Then, inspired by the inception module~\cite{szegedy2016inception}, we treat $\hat{\mathbf{A}}_m$ as convolution kernel and $\widetilde{\mathbf{X}}^{(0)}$ as an initial residual term carrying non-smoothed features to encode multi-scale structural information based on the Eq.~(\ref{eq:magnetic_feature_aggre}).} 

    In a nutshell, LightDiC first executes $K$-step feature smoothing by the low-frequency spectrum of $\mathbf{L}_m$. 
    Then, it leverages these smoothed features to encode multi-scale structural insights $\widetilde{\mathbf{X}}$.
    We notice that a recent work SIGN~\cite{frasca2020sign} uses a learnable combination. 
    However, as we previously pointed out, naive weighted message aggregation is not applicable to intricate processes in the complex domain. 
    We will elaborate on this matter extensively in~\cite{LightDic}.

\noindent\textbf{Model training.}
    To eliminate the unnecessary complexity, we fold the complex learning function into a linear predictor.
     {It efficiently reduces the computational overhead associated with recursive calculations involving the complete square expansions of real and imaginary components. 
    For large-scale learning grounded in the complex domain, the benefits of this simplicity are self-evident.}  
    \begin{equation}
    \label{eq:s2diconv}
        \begin{aligned}
    \mathbf{Z}=\mathrm{Softmax}\left(\left(\text{Real}\left(\widetilde{\mathbf{X}}\right)||\text{Imag}\left(\widetilde{\mathbf{X}}\right)\right)\mathbf{W}\right),
        \end{aligned}
    \end{equation}
    where $\cdot||\cdot$ represents the concat operation. 
    To perform link-level tasks, we concatenate the embeddings of node pairs for prediction.
    Notably, complex model architectures and more learnable parameters may yield higher predictive performance. 
    However, the purpose of our model framework is to highlight the validity and scalability of complex domain-based feature propagation.

\subsection{Theorem Analysis}
\label{sec: Theorem Analysis}
\textbf{Smoothness and Dirichlet energy on digraphs.}
   We first define the smoothness of digraphs based on the magnetic Laplacian and Euler equation. 
    Specifically, we can integrate the angle $\theta_{uv}$ associated with each edge $e_{uv}$ and have Definition~\ref{def: smoothness} and Lemma~\ref{lemma:xLx}.

\begin{definition}
\label{def: smoothness}
    Let $\mathbf{L}_m\in\mathbb{C}^{N\times N}$ be the magnetic Laplacian of a digraph $\mathcal{G}$. 
    Given node feature matrix $\mathbf{X}\in\mathbb{C}^{N\times 1}$, the complex domain-based smoothness of $\mathbf{X}$ over $\mathcal{G}$ is defined as $\mathbf{X}^\dagger\mathbf{L}_m\mathbf{X}$.
\end{definition}

\begin{lemma}
\label{lemma:xLx}
    The total variation of the digraph signal $\mathbf{X}$ is a smoothness measure, quantifying how much the signal $\mathbf{X}$ changes with respect to the digraph topology encoded in magnetic Laplacian $\mathbf{L}_m$ as the following quadratic form, which is also known as Dirichlet energy 
\begin{equation}
\label{eq: smoothness in digraphs}
    \begin{aligned}
    \mathbf{X}^\dagger\mathbf{L}_m\mathbf{X}=\!\!\!\sum_{(u,v)\in \mathcal{E}}\!\!\!{|\mathbf{X}_u-e^{i\theta_{uv}}\mathbf{X}_v|^2}=\!\!\!\sum_{(u,v)\in  \mathcal{E}}\!\!\!{|e^{-i\theta_{uv}}\mathbf{X}_u-\mathbf{X}_v|^2}.
    \end{aligned}
\end{equation}
\end{lemma}
    Then, from the perspective of signal denoising, suppose that the digraph signal $\mathbf{y}$ has noises ${\epsilon}$, we can define the objective function that minimizes the global Dirichlet energy of digraphs as follows
\begin{definition}
\label{def: graph_signal_denoise}
    Given digraph signal $\mathbf{y}\in\mathbb{C}^{N\times 1}$ with noises ${\epsilon}$, the optimization function of global Dirichlet energy is defined as
\begin{equation}
\label{eq: graph_signal_denoise}
    \begin{aligned}
    \min_{\mathbf{X}\in\mathbb{C}}\mathbf{Z}(\mathbf{x})=\min_{\mathbf{X}\in\mathbb{C}}\Vert\mathbf{X}-\mathbf{y}\Vert_2+\mathbf{X}^\dagger\mathbf{L}_m\mathbf{X}.
    \end{aligned}
\end{equation}
\end{definition} 
    Here $\min_{\mathbf{X}\in\mathbb{C}}\mathbf{Z}(\mathbf{x})$ consists of two terms: $\mathbf{X}^\dagger\mathbf{L}_m\mathbf{X}$ measures the smoothness of resulting signals, and $\Vert\mathbf{X}-\mathbf{y}\Vert_2$ guarantees that the resulting signals $\mathbf{y}$ keep information of the original signals $\mathbf{X}$. 

\noindent\textbf{Spectral analysis of complex domain propagation.}
    Based on the Eq.~(\ref{eq: graph_signal_denoise}), it is known that $\mathbf{X}^*=(\mathbf{L}_m+\mathbf{I})^{-1}\mathbf{y}$ is the solution of this optimization. 
    However, directly computing the inverse of $\mathbf{L}_m+\mathbf{I}$ is prohibitive on large-scale scenarios, 
    LightDiC implicitly optimizes it through the spectral analysis of magnetic Laplacian.
    To demonstrate our point, we first prove that LightDiC is essentially a low-pass filter (Lemma~\ref{lemma: low-pass}) and the small eigenvalues are referred to relatively smooth eigenvectors in the eigendecomposition (Lemma~\ref{lemma: low-eigen}).
\begin{lemma}
\label{lemma: low-pass}
    In digraph signal processing perspective, the complex domain feature propagation in our proposed LightDiC is a digraph convolution operation $\hat{\mathbf{A}}_m^{K}$ with a low-pass filter $g(\lambda_i)=(1-\lambda_i)^K$.
\end{lemma}
    Similar to the GCN, we fit the convolution kernel using a first-order approximation of Chebyshev polynomials, which retains eigenvector components related to lower eigenvalues and discards those related to higher eigenvalues.
    Here we further investigate what the low eigenvalues of the digraph reflect.
\begin{lemma}
\label{lemma: low-eigen}
    Let $\mathbf{L}_m\in\mathbb{C}^{n\times n}$ be a Hermitian matrix with eigenvalues $\lambda_1\leq\dots\leq\lambda_n$, we have $\lambda_k=\max_{S\in\mathbb{C}^n,\ dim(S)=k}\min_{\mathbf{X}\in S}\frac{ \mathbf{X}^\dagger\mathbf{L}_m\mathbf{X}}{\mathbf{X}^\dagger\mathbf{X}}$ and $ \lambda_1=\min_{\mathbf{X}\in\mathbb{C}^n}{\mathbf{X}^\dagger\mathbf{L}_m\mathbf{X}}$. 
    The smoothness $\mathbf{X}^\dagger\mathbf{L}_m\mathbf{X}$ is minimized by the eigenvector $\mathbf{u}_1$ corresponding to the smallest eigenvalue $\lambda_1$.
\end{lemma}
    Building upon this, we can derive that the smaller eigenvalues have corresponding eigenvectors with bigger smoothness.
    In other words, the magnetic Laplacian-guided complex domain message passing (i.e., feature pre-process in LightDiC) is essentially a process of smoothing digraph nodes based on the low eigenvalue spectrum.

\noindent\textbf{LightDiC and Dirichlet energy optimization function.}
    Finally, as we report in \cite{LightDic}, LightDiC implicitly optimizes the digraph smoothness, which corresponds to the proximal gradient descent process of the Dirichlet energy optimization function as follows
\begin{lemma}
\label{lemma: pgd}
    Given the digraph Fourier transform for a signal $\mathbf{X}: \mathbb{V} \rightarrow \mathbb{C}$ by $\hat{\mathbf{X}}=\mathbf{U}^{\dagger} \mathbf{X}$, so that $\widehat{\mathbf{X}}(k)=\left\langle\mathbf{X}, \mathbf{u}_k\right\rangle$, the unitary complex numbers $\mathbf{u}_1, \cdots, \mathbf{u}_n$ as the Fourier basis for digraphs, the Fourier inverse formula is $\mathbf{X}=\mathbf{U} \widehat{\mathbf{X}}=\sum_{k=1}^n \widehat{\mathbf{X}}(k) \mathbf{u}_k$.
    Start from $\widetilde{\mathbf{X}}^{(0)}=\mathbf{X}$ (node feature), $\hat{\mathbf{A}}_m^{K}\mathbf{X}$ implicitly optimize the Fourier inverse formula by applying $K$ proximal gradient descent steps.
\end{lemma}
    Until now, we have provided the theoretical generalization of spectral GNNs to digraphs through the magnetic Laplacian. 
    While the magnetic Laplacian has garnered significant attention in graph theory, its integration with DiGNNs is both novel and essential. 
    This integration provides a unified theoretical framework (i.e. Smoothness, Dirichlet energy, and Spectral analysis) and model design principles, which explain why LightDiC can achieve impressive performance on a simple linear predictor.
    Simultaneously, the interpretability and transparent nature of this approach are crucial for instilling confidence in applying LightDiC to real-world scenarios.

\begin{table*}[t]
\setlength{\abovecaptionskip}{0.2cm}
\caption{The statistical information of the experimental digraph datasets.
}
\footnotesize 
\label{tab: datasets}
\resizebox{\linewidth}{15mm}{
\setlength{\tabcolsep}{0.9mm}{
\begin{tabular}{ccccccccc}
\midrule[0.3pt]
Datasets        & \#Node      & \#Edges       & \#Features & \#Node Classes & \#Node Train/Val/Test & \#Edge Train/Val/Test & \#Task     & Description       \\ \midrule[0.3pt]
CoraML          & 2,995       & 8,416         & 2,879      & 7              & 140/500/2355          & 80\%/15\%/5\%         & Node/Link-level & citation network  \\
CiteSeer        & 3,312       & 4,591         & 3,703      & 6              & 120/500/2692          & 80\%/15\%/5\%         & Node/Link-level & citation network  \\
WikiCS          & 11,701      & 290,519       & 300        & 10             & 580/1769/5847         & 80\%/15\%/5\%         & Node/Link-level & weblink network   \\
Slashdot        & 75,144      & 425,702       & 100        & -              & -                     & 80\%/15\%/5\%         & Link-level            & social network    \\
Epinions        & 114,467     & 717,129       & 100        & -              & -                     & 80\%/15\%/5\%         & Link-level            & social network    \\
WikiTalk        & 2,388,953   & 5,018,445     & 100        & -              & -                     & 80\%/15\%/5\%         & Link-level            & co-editor network \\
ogbn-papers100M & 111,059,956 & 1,615,685,872 & 128        & 172            & 1207k/125k/214k       & 80\%/15\%/5\%         & Node/Link-level & citation network  \\ \midrule[0.3pt]
\end{tabular}
}}
\end{table*}
\section{Experiments}
\label{sec: Experiments}
    In this section, we first introduce experimental setups. 
    Then, we aim to answer the following questions to verify the effectiveness of our proposed LightDiC: 
    \textbf{Q1}: Compared to both existing DiGNNs and undirected scalable GNNs, how does LightDiC perform in terms of predictive accuracy and efficiency?
    \textbf{Q2}: If LightDiC is effective, what contributes to its performance gain? 
    \textbf{Q3}: How does LightDiC perform when applied to real-world sparse digraphs?

\subsection{Experimental Setup}
\textbf{Datasets.}
    Citation networks (CoraML, Citeseer, and papers100M) in~\cite{bojchevski2018coraml_citeseer,hu2020ogb}, social networks (Slashdot and Epinions) in~\cite{ordozgoiti2020slashdot,massa2005epinions}, web-link network (WikiCS) in ~\cite{mernyei2020wikics}, and co-editor network (WikiTalk) in~\cite{leskovec2010wikitalk}.
    The dataset statistics are shown in Table.\ref{tab: datasets} and more descriptions can be found in~\cite{LightDic}.
    Notably, we use the directed version of datasets instead of the one provided by the PyG library (CoraML, CiteSeer)\footnote{https://pytorch-geometric.readthedocs.io/en/latest/modules/datasets.html}, WikiCS paper\footnote{https://github.com/pmernyei/wiki-cs-dataset} and the raw data given by the OGB (ogb-papers100M)\footnote{https://ogb.stanford.edu/docs/nodeprop/}.
    In addition, for Slashdot, Epinions, and WikiTalk, the PyGSD~\cite{he2022pygsd_library} library reveals only the topology and lacks the corresponding node features and labels.
    Therefore, we generate the node features using eigenvectors of the regularised topology.

\noindent\textbf{Baselines.} 
     {
    (1) Spatial: DGCN~\cite{tong2020dgcn}, DIMPA~\cite{he2022dimpa} and NSTE~\cite{kollias2022nste}; 
    (2) Mix: DiGCN~\cite{tong2020digcn} and its two variants, DiGCN-Appr, DiGCN-IB; 
    (3) Spectral: MagNet~\cite{zhang2021magnet} and MGC\cite{zhang2021mgc}.
    To verify the scalability and generalization of LightDiC, we compare with the undirected baselines: GCN~\cite{kipf2016gcn}, GraphSAGE~~\cite{hamilton2017graphsage}, UniMP~\cite{shi2020unimp}, SGC~\cite{wu2019sgc}, SIGN~\cite{frasca2020sign}, GBP~\cite{chen2020gbp}, S$^2$GC~\cite{zhu2021ssgc}, and GAMLP~\cite{gamlp}. 
    The descriptions of them are provided in~\cite{LightDic}.
    For link-level dataset split, we are aligned with previous work~\cite{zhang2021magnet,he2022msgnn,he2022pygsd_library}. 
    We repeat each experiment 10 times to represent unbiased performance and running time (second report).
    We also employ multiple metrics AUC, Macro-F1, and Accuracy to evaluate experimental results, with Accuracy as the default metric.}

\begin{table*}[t]
\centering
\setlength{\abovecaptionskip}{0.2cm}
\setlength{\belowcaptionskip}{-0.2cm}
\caption{ {Performance, pre-process/training/inference time and learnable parameters.
The best result is \textbf{bold}.
The second result is \underline{underline}.
}}
\footnotesize 
\label{tab: NC_cmp}
\resizebox{\linewidth}{20mm}{
\setlength{\tabcolsep}{1mm}{
\begin{tabular}{c|cccc|cccc|cccc}
\hline
\multirow{2}{*}{Node Classification} & \multicolumn{4}{c|}{CoraML}                                            & \multicolumn{4}{c|}{CiteSeer}                                          & \multicolumn{4}{c}{WikiCS}                                            \\
                            & Test Acc            & (Pre.)Train             & Infer.         & Param.        & Test Acc            & (Pre.)Train             & Infer.         & Param.        & Test Acc            & (Pre.)Train             & Infer.         & Param.       \\ \hline
DGCN                        & 80.35±0.83          & (2.4) 11.3          & 0.36          & 200K         & 61.16±1.32          & (2.1) 14.6          & 0.42          & 253K         & 78.25±0.61          & (4.7) 84.8          & 0.63          & 37K         \\
DiGCN                       & 80.72±0.92          & (2.0) 15.4          & 0.42          & 580K         & 62.70±1.05          & (2.2) 19.5          & 0.50          & 738K         & 79.73±0.57          & (4.4) 28.4          & 0.45          & 86K         \\
DiGCN-IB                    & 80.86±0.90          & (2.8) 17.5          & 0.47          & 580K         & 62.78±1.22          & (2.6) 21.2          & 0.54          & 738K         & \textbf{80.05±0.54} & (5.8) 159.3         & 0.70          & 86K         \\
DiGCN-Appr                  & 80.74±0.31          & \underline{(1.4) {4.36}}& \underline{0.08} & \underline{190K} & 60.57±0.60  & \underline{(1.3) {5.50}} & \underline{0.12}  & \underline{242K}         & 79.31±0.34          & \underline{(2.8)} 8.35          & 0.18          & \underline{25K}         \\
DIMPA                       & 81.12±0.84          & (-) 11.8          & 0.36          & 371K         & 61.64±1.25          &  (-) 14.2         & 0.45          & 476K         & 78.88±0.42          & (-) 47.0          & 0.55          & 42K         \\
NSTE                        & 81.75±0.96          & (2.5) 9.81          & 0.30          & 370K         & 61.58±1.59          & (2.8) 11.9          & 0.39          & 475K         & 79.05±0.53          & (5.0) 25.4           & 0.46          & 40K         \\
MagNet                      & 81.48±0.70          & (\textbf{1.0}) 11.4          & 0.18          & 380K         & \underline{63.46±1.04}          & (\textbf{0.7}) 11.9          & 0.27          & 485K         & 79.59±0.39          & (\textbf{1.5}) 14.6          & 0.30          & 51K         \\
MGC                         & \underline{84.08±0.94} & (2.2) 4.60          & 0.10          & \underline{190K}         & 63.25±1.35          & (2.0) 7.27          & 0.15          & \underline{242K}         & 79.26±0.48          & (5.4) \underline{8.12}          & \underline{0.08}          & \underline{25K}         \\
LightDiC (ours)                    & \textbf{84.16±0.72}          &(1.6) \textbf{1.95} & \textbf{0.04} & \textbf{40K} & \textbf{63.74±0.81} & (1.5) \textbf{4.30} & \textbf{0.06} & \textbf{45K} & \underline{79.84±0.36}          & (3.6) \textbf{5.60} & {0.03} & \textbf{6K} \\ \hline
\end{tabular}
}}
\end{table*}

\begin{table*}[]
\setlength{\abovecaptionskip}{0.2cm}
\setlength{\belowcaptionskip}{-0.2cm}
\caption{ {Performance under suitable topology for each method.
OOM and OOT are the out-of-memory and more than 2 hours of training.}
}
\label{tab: link_cmp}
\resizebox{\linewidth}{42mm}{
\setlength{\tabcolsep}{1.8mm}{
\begin{tabular}{cc|cccccccccc}
\midrule[0.3pt]
Datasets                                                                                       & Tasks     & GCN      & SAGE     & SIGN     & GAMLP    & DiGCN    & DIMPA    & NSTE     & MGC      & MagNet   & LightDiC \\ \midrule[0.3pt]
\multirow{3}{*}{\begin{tabular}[c]{@{}c@{}}CoraML\\ E-AUC\\ D-Macro-F1\end{tabular}}   & Existence & 84.5±0.2 & 85.9±0.3 & 85.6±0.2 & 86.0±0.3 & 87.9±0.3 & 88.4±0.5 & 88.7±0.4 & \textbf{89.4±0.3} & 88.6±0.4 & \underline{89.2±0.2} \\
                                                                                               & Direction & 82.6±0.3 & 82.3±0.4 & 83.1±0.3 & 83.8±0.4 & 86.7±0.3 & 87.5±0.3 & 87.6±0.4 & \underline{87.9±0.2} & 87.5±0.3 & \textbf{88.0±0.2} \\
                                                                                               & Link-C    & 69.2±0.4 & 69.4±0.5 & 68.9±0.4 & 69.3±0.5 & 72.9±0.5 & 74.0±0.4 & \underline{74.2±0.6} & {74.0±0.2} & 73.8±0.4 & \textbf{74.8±0.3} \\ \midrule[0.3pt]
\multirow{3}{*}{\begin{tabular}[c]{@{}c@{}}CiteSeer\\ E-AUC\\ D-Macro-F1\end{tabular}} & Existence & 76.8±0.2 & 77.3±0.3 & 78.2±0.3 & 78.8±0.4 & 84.6±0.3 & 84.8±0.4 & 84.7±0.5 & 84.9±0.3 & 
\underline{85.3±0.4} & \textbf{86.1±0.3} \\
                                                                                               & Direction & 79.2±0.4 & 79.1±0.3 & 79.4±0.3 & 79.7±0.5 & 84.7±0.4 & 85.2±0.3 & 84.8±0.4 & \underline{85.6±0.3} & 85.5±0.3 & \textbf{86.8±0.2} \\
                                                                                               & Link-C    & 62.3±0.4 & 62.6±0.5 & 62.4±0.4 & 62.5±0.6 & 63.8±0.4 & 64.0±0.5 & \underline{64.3±0.6} & 64.1±0.3 & 64.0±0.3 & \textbf{65.2±0.2} \\ \midrule[0.3pt]
\multirow{3}{*}{\begin{tabular}[c]{@{}c@{}}WikiCS\\ E-AUC\\ D-Macro-F1\end{tabular}}   & Existence & 86.9±0.2 & 87.2±0.2 & 86.9±0.3 & 87.1±0.4 & \underline{89.6±0.3} & 89.4±0.2 & \textbf{89.7±0.2} & 89.5±0.1 & 89.5±0.2 & \underline{89.6±0.2} \\
                                                                                               & Direction & 84.6±0.2 & 85.0±0.3 & 84.8±0.3 & 85.2±0.3 & 87.6±0.4 & \textbf{87.8±0.4} & 87.5±0.3 & 87.6±0.2 & 87.6±0.2 & \underline{87.7±0.1} \\
                                                                                               & Link-C    & 75.2±0.3 & 75.6±0.2 & 75.4±0.3 & 75.7±0.4 & \underline{80.0±0.3} & 79.8±0.3 & 79.8±0.4 & 79.6±0.2 & {79.9±0.1} & \textbf{80.4±0.2} \\ \midrule[0.3pt]
\multirow{3}{*}{\begin{tabular}[c]{@{}c@{}}Slashdot\\ E-AUC\\ D-Macro-F1\end{tabular}} & Existence & 87.3±0.3 & 87.6±0.5 & 87.9±0.5 & 88.0±0.4 & 90.3±0.3 & 90.1±0.4 & 90.3±0.3 & \underline{90.5±0.1} & 90.4±0.2 & \textbf{90.8±0.2} \\
                                                                                               & Direction & 85.4±0.3 & 85.6±0.3 & 85.8±0.3 & 86.0±0.4 & 90.4±0.2 & \textbf{90.6±0.1} & \underline{90.5±0.1} & {90.3±0.1} & 90.4±0.1 & \underline{90.5±0.1} \\
                                                                                               & Link-C    & 78.4±0.2 & 77.8±0.1 & 78.5±0.4 & 78.5±0.3 & 84.1±0.1 & \underline{84.3±0.1} & 84.1±0.2 & 84.0±0.1 & 84.1±0.2 & \textbf{84.5±0.1} \\ \midrule[0.3pt]
\multirow{3}{*}{\begin{tabular}[c]{@{}c@{}}Epinions\\ E-AUC\\ D-Macro-F1\end{tabular}} & Existence & 89.4±0.1 & 89.0±0.1 & 89.2±0.2 & 89.5±0.2 & \underline{94.2±0.1} & 94.0±0.2 & 93.8±0.1 & \underline{94.2±0.1} & 94.0±0.1 & \textbf{94.4±0.1} \\
                                                                                               & Direction & 83.3±0.2 & 83.5±0.2 & 83.4±0.2 & 83.7±0.1 & 85.9±0.1 & \underline{86.0±0.1} & 85.8±0.1 & {85.8±0.2} & \textbf{86.2±0.1} & \underline{86.0±0.1} \\
                                                                                               & Link-C    & 82.2±0.2 & 82.4±0.1 & 81.9±0.1 & 82.5±0.1 & 85.7±0.1 & 86.1±0.1 & 85.9±0.2 & 86.2±0.1 & \textbf{86.5±0.2} & \underline{86.4±0.1} \\ \midrule[0.3pt]
\multirow{3}{*}{\begin{tabular}[c]{@{}c@{}}WikiTalk\\ E-AUC\\ D-Macro-F1\end{tabular}} & Existence & 90.2±0.1 & 90.0±0.1 & 90.3±0.1 & 90.3±0.1 & OOT      & OOM      & 94.6±0.1 & \underline{94.8±0.1} & OOM      & \textbf{95.3±0.1} \\
                                                                                               & Direction & 86.5±0.1 & 86.4±0.1 & 86.5±0.1 & 86.7±0.1 & OOT      & OOM      & \underline{91.5±0.1} & 91.4±0.1 & OOM      & \textbf{91.8±0.1} \\
                                                                                               & Link-C    & 85.2±0.1 & 85.6±0.1 & 85.4±0.1 & 85.5±0.1 & OOT      & OOM      & \underline{90.2±0.1} & 90.1±0.1 & OOM      & \textbf{90.5±0.1} \\ \midrule[0.3pt]
\end{tabular}
}}
\end{table*}

\begin{table}[]
\setlength{\abovecaptionskip}{0.2cm}
\setlength{\belowcaptionskip}{-0.2cm}
\caption{Performance on directed ogbn-papers100M within 12 hours. 
}
\label{tab: papers100m_cmp}
\resizebox{\linewidth}{28mm}{
\setlength{\tabcolsep}{1mm}{
\begin{tabular}{cc|cccc}
\midrule[0.3pt]
Type                        & Model             & Node-C                    & Existence                 & Direction                 & Link-C            \\ \midrule[0.3pt]
\multirow{8}{*}{Undirected} & MLP               & 47.2±0.3                  & 86.5±0.1                  & 90.4±0.1                  & 85.6±0.2          \\
                            & SGC               & 45.8±0.1                  & 84.6±0.1                  & 87.6±0.1                  & 83.1±0.1          \\
                            & GBP               & 48.3±0.2                  & 85.3±0.2                  & 88.4±0.1                  & 84.5±0.1          \\
                            & SIGN              & 52.5±0.2                  & 86.8±0.2                  & 89.5±0.2                  & 85.0±0.1          \\
                            & S$^2$GC           & 50.6±0.1                  & 86.4±0.1                  & 88.5±0.2                  & 84.6±0.1          \\
                            & SAGE              & 55.2±0.2                  & 87.4±0.1                  & \underline{91.0±0.2}      & 86.1±0.2            \\
                            & UniMP             & 54.7±0.3                  & {87.3±0.2}                & 90.2±0.2                  & \underline{86.4±0.2}   \\
                            & GAMLP             & \underline{56.8±0.3}      & \underline{87.7±0.2}      & {90.6±0.1}                & 86.0±0.2         \\ \midrule[0.3pt]
\multirow{3}{*}{Directed}   & MGC               & OOT                       & OOT                       & OOT                       & OOT               \\
                            & NSTE              & OOM                       & OOM                       & OOM                       & OOM               \\
                            & MagNet            & OOM                       & OOM                       & OOM                       & OOM               \\
                            & {LightDiC}        & \textbf{65.4±0.2}         & \textbf{91.6±0.1}         & \textbf{93.8±0.1}         & \textbf{90.3±0.1} \\ \midrule[0.3pt]
\end{tabular}
}}
\end{table}

\noindent\textbf{Implementation Details.}
    The hyper-parameters of baselines are tuned by Optuna~\cite{akiba2019optuna} or set according to the original paper if available.
     {The parameter $q$ and $K$ are acquired by means of interval search from $\{0, 0.25\}$ and $\{2, 10\}$. 
    As for the learning rate, we use a grid search from $\{0.1,0.001\}$.
    We employ mini-batch training with batch size 5K by default.
    Moreover, the experiments are conducted on the machine with Intel(R) Xeon(R) Gold 6230R CPU @ 2.10GHz, and NVIDIA GeForce RTX 3090 with 24GB memory and CUDA 11.8.
    The operating system is Ubuntu 18.04.6 with 768GB memory.}

\subsection{End-to-end Comparison}
    \textbf{Node-level Performance.}
    To answer \textbf{Q1}, we report the node-level performance in Table~\ref{tab: NC_cmp} and Table~\ref{tab: papers100m_cmp}. 
    Our findings indicate that LightDiC consistently achieves either the highest or second-highest performance across CoraML, CiteSeer, and WikiCS. 
    This is attributed to the explicit digraph signal smoothing mechanism as proved in~\cite{LightDic}, setting it apart from MagNet.
    Moreover, DiGCN, DIMPA, and NSTE exhibit instability and slightly worse performance due to over-fitting.
    For the large-scale ogbn-papers100M, we observe that existing DiGNNs lack the necessary scalability, resulting in OOM and OOT errors. 
     {For undirected baselines, significant performance degradation occurs both in spectral and spatial methods. 
    In the former, the lack of a theoretical foundation supporting the spectral analysis of asymmetric topologies leads to abnormal eigenvalues, misleading the model and adversely affecting predictive performance. 
    In the latter, adherence to a strictly spatially symmetric message-passing mechanism hinders the recognition of intricate directed relationships, resulting in sub-optimal performance.
    For more extensive performance analysis, please refer to Sec.~\ref{sec: Efficiency and Scalability Analysis}.}

    \noindent\textbf{Link-level Performance.}
     {Table~\ref{tab: link_cmp} and Table~\ref{tab: NC_cmp} show that LightDiC achieves high scalability and exhibits best performance on CiteSeer and WikiTalk in all three link-level tasks with various evaluation metrics.} 
    Even without achieving the best performance, LightDiC ranks as a powerful contender for the second-best performance as other methods lack uniform competitiveness. 
    The remarkable performance in link-level tasks signifies its adeptness in capturing complex directed topologies. 
    Moreover, we observe that raw features are better suited for direction prediction, while topology-based generated features are more effective for predicting existence.
     {Notably, undirected GNNs under coarse undirected transformation do not present satisfying results in three link-level tasks, which intuitively validates their inability to cope with directed topologies as they struggle to capture rich interactions among nodes and extract knowledge from information-impaired undirected representations.}

\subsection{Efficiency and Scalability Analysis}
\label{sec: Efficiency and Scalability Analysis}
\begin{figure*}[t]   
	\centering
    \setlength{\abovecaptionskip}{0.2cm}
	\setlength{\belowcaptionskip}{-0.2cm}
 \includegraphics[width=\linewidth,scale=1.00]{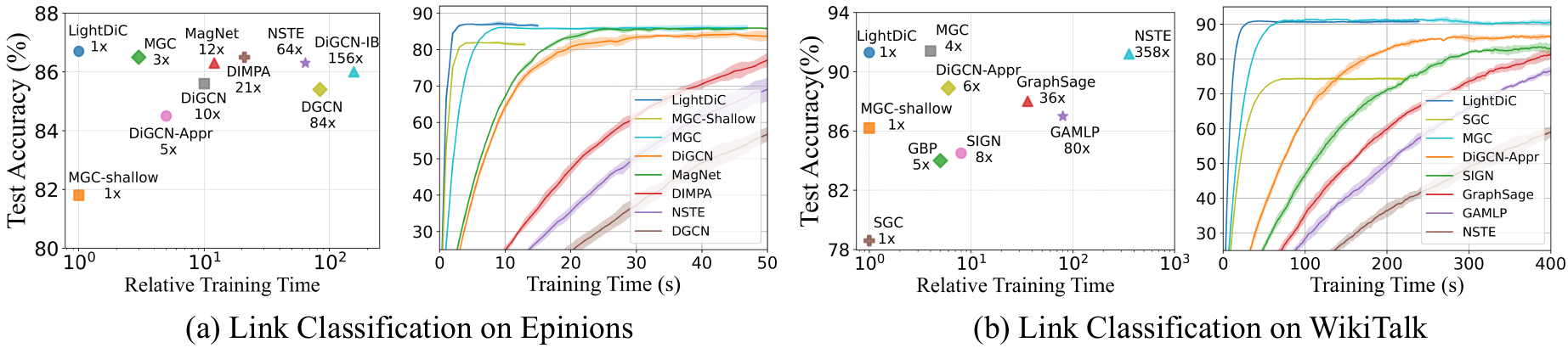}
	\caption{
    Convergence curves with the relative training time on Epinions and WikiTalk datasets. 
    The shaded area is the result range of 10 runs.}
	\label{fig:scalability}
\end{figure*}

     {To answer \textbf{Q1}, we provide efficiency reports in Table~\ref{tab: NC_cmp}, Table~\ref{tab: papers100m_run_efficienc}, and Fig.~\ref{fig:scalability}.
    Notably, in the pre-process, DGCN and NSTE generate high-order node dependency to increase RF. 
    DiGCN generates a symmetric digraph Laplacian, while DIMPA directly increases RF during training.
    In Table~\ref{tab: NC_cmp}, LightDiC achieves remarkable performance while significantly reducing running time and parameters, resulting in gains of up to 30x and 16x.
    Meanwhile, LightDiC maintains stable and competitive performance in Table~\ref{tab: link_cmp} while DiGCN, DIMPA, and MagNet encounter OOM and OOT errors. 
    Moreover, Table~\ref{tab: papers100m_cmp} reveals that all the existing DiGNNs encounter failures.} 
    
     {According to the Table~\ref{tab: papers100m_cmp} and Table~\ref{tab: papers100m_run_efficienc}, our findings are as follows:
    (1) Relying on more complex architectures, SIGN and GAMLP incur increased complexity but with better results in comparison. 
    Yet they remain less competitive in directed scenarios. 
    In contrast, LightDiC introduces computations in the complex domain, having slightly higher complexity than SGC. 
    But this trade-off yields satisfactory performance.
    (2) While without pre-processing, GraphSAGE and UniMP introduce additional trainable weights and sampling processes that must be executed in every epoch, which leads to additional memory costs and significant computational complexity.}
    
     {We observe that most weight-free pre-processing computations can be represented as sparse matrix multiplications. 
    They are easily parallelism through distributed frameworks and accelerated by tailored matrix computing strategies, allowing us to focus more on training, inference, and memory costs. 
    Considering the predictive performance shown in Table~\ref{tab: NC_cmp}, Table~\ref{tab: link_cmp}, and Table~\ref{tab: papers100m_cmp}, LightDiC exhibits significant advantages in these three complexity aspects.}

    To further validate scalability, we provide a visualization of the training efficiency in Fig.~\ref{fig:scalability}, where MGC-Shallow represents the low computational overhead variant of MGC with shallow filter order and propagation steps. 
    It demonstrates that LightDiC maintains impressive scalability and performance with low complexity, as evidenced by quick convergence and high efficiency.
    For a more comprehensive analysis of computational and storage efficiency from an algorithmic complexity perspective, please refer to Sec.~\ref{sec: Complexity Analysis}.

\begin{figure}[t]   
	\centering
    \setlength{\abovecaptionskip}{0.1cm}
    \setlength{\belowcaptionskip}{-0.1cm}
	\includegraphics[width=\linewidth,scale=1.00]{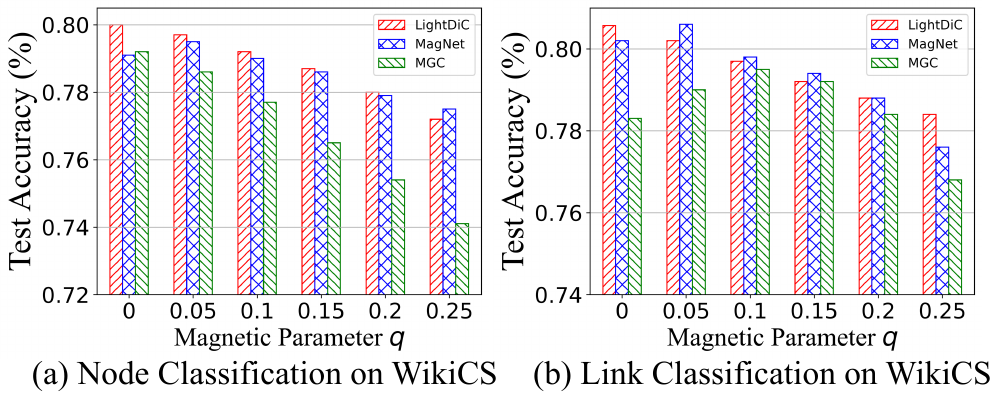}
	\caption{
    Performance of magnetic Laplacian-based DiGNNs.}
	\label{fig:q}
\end{figure}

\subsection{Sensitivity Analysis}
     {To answer \textbf{Q2}, we delve into two pivotal factors influencing the performance of LightDiC: the number of propagation steps $L$ in the pre-process and the magnetic parameter $q$ in the training phase.
    Table~\ref{tab: model_layer} reveals that MGC's reliance on approximated linear rank necessitates the involvement of deep propagated features,
    which hinders its scalability(i.e., OOT errors in Table~\ref{tab: papers100m_cmp}). 
    Meanwhile, the coupled architecture of MagNet limits its potential for deep design in medium- and large-scale digraph datasets.} 
    In contrast, LightDiC benefits from flexibility in its decouple and simple learning architecture, while both MagNet and LightDiC encounter over-smoothing challenges.
    Additionally, Fig.~\ref{fig:q} illustrates that the model performance hinges on the appropriate choice of $q$, which governs the angle of complex-domain feature propagation between nodes. 
     {The deep architecture of MGC exacerbates the impacts of unsuited $q$.
    In contrast, both LightDiC and MagNet exhibit a more consistent performance across a range of $q$ values due to appropriate propagation steps, signifying their stability in this regard.}


\begin{table}[]
\setlength{\abovecaptionskip}{0.2cm}
\setlength{\belowcaptionskip}{-0.2cm}
\caption{ {Epoch-Batch efficiency on directed ogbn-papers100M.}
}
\label{tab: papers100m_run_efficienc}
\resizebox{\linewidth}{20mm}{
\setlength{\tabcolsep}{1mm}{
\begin{tabular}{cccccc}
\midrule[0.3pt]
Method   & Pre-process(s)  & E-Train.(s)   & E-Infer.(s)   & B-GPU Mem. & Param. \\ \midrule[0.3pt]
MLP      & -            & 8.80±0.28  & 4.42±0.22   & 8233M     & 151K   \\
SGC      & 854.33±1.56  & 8.24±0.22  & 4.36±0.22   & 2897M      & 22K   \\
GBP      & 735.60±1.22  & 9.26±0.25  & 4.90±0.19   & 9476M     & 151K    \\
S$^2$GC  & 976.83±1.17 & 9.45±0.26  & 6.14±0.33   & 10431M     & 151K   \\
SIGN     & 1062.06±0.86  & 11.72±0.31 & 6.77±0.49   & 10760M     & 165K    \\
SAGE     & -            & 47.94±1.10 & 122.31±1.18 & 16392M     & 342K   \\
UniMP    & -            & 64.13±1.46 & 118.85±1.50 & 22764M     & 659K   \\
GAMLP    & 1384.76±1.60 & 13.68±0.40 & 8.10±0.43   & 16378M     & 455K   \\ \midrule[0.3pt]
LightDiC & 945.72±1.43  & 8.75±0.18  & 4.47±0.24   & 4269M     & 45K   \\ \midrule[0.3pt]
\end{tabular}
}}
\end{table}

\begin{table}[]
\setlength{\abovecaptionskip}{0.2cm}
\setlength{\belowcaptionskip}{-0.2cm}
\caption{Performance on WikiCS with different model depths.
}
\label{tab: model_layer}
\resizebox{\linewidth}{21mm}{
\setlength{\tabcolsep}{1mm}{
\begin{tabular}{cc|cccc}
\midrule[0.3pt]
Tasks                   & Model Depth & NSTE              & MGC               & MagNet            & LightDiC          \\ \midrule[0.3pt]
\multirow{4}{*}{Node-C} & 2-Layer     & \textbf{79.1±0.5} & 72.5±1.4          & \textbf{79.6±0.4} & \textbf{79.8±0.4} \\
                        & 8-Layer     & OOM               & 76.6±1.0          & 73.7±0.3          & 78.8±0.3          \\
                        & 32-Layer    & OOM               & \textbf{79.3±0.5} & OOM               & 77.4±0.3          \\
                        & 64-Layer    & OOM               & 78.5±0.6          & OOM               & 76.3±0.2          \\ \midrule[0.3pt]
\multirow{4}{*}{Link-C} & 2-Layer     & \textbf{79.8±0.4} & 75.8±0.9          & \textbf{80.0±0.1} & \textbf{80.2±0.2} \\
                        & 8-Layer     & OOM               & 78.5±0.7          & 78.3±0.2          & 79.9±0.1          \\
                        & 32-Layer    & OOM               & \textbf{79.6±0.2} & OOM               & 78.9±0.2          \\
                        & 64-Layer    & OOM               & 79.0±0.1          & OOM               & 78.2±0.1          \\ \midrule[0.3pt]
\end{tabular}
}}
\end{table}

\begin{figure}[t]   
	\centering
    \setlength{\abovecaptionskip}{0.2cm}
    \setlength{\belowcaptionskip}{-0.1cm}
	\includegraphics[width=\linewidth,scale=1.00]{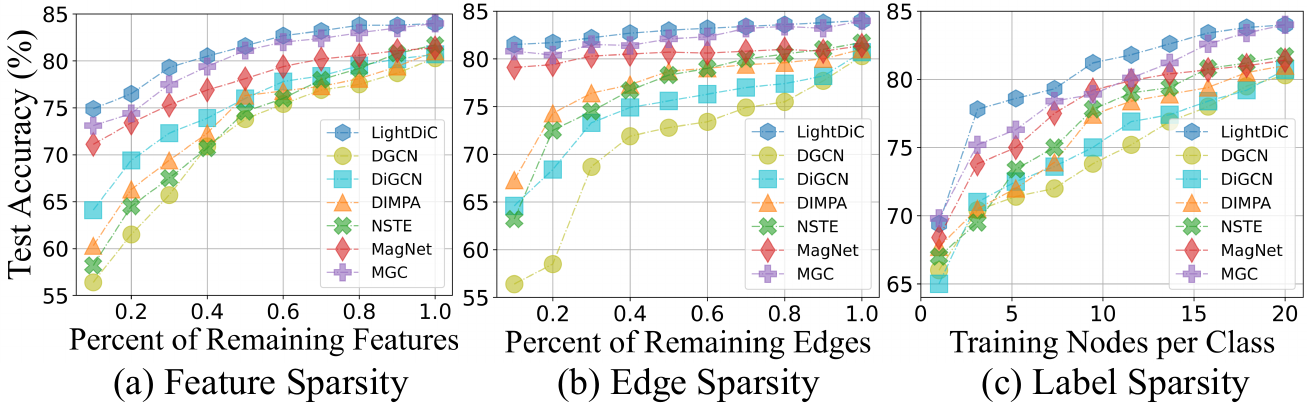}
	\caption{
    Node-C performance on CoraML under sparsity settings.}
	\label{fig:sparsity}
\end{figure}

\subsection{Performance on Sparse Digraphs} 
    To answer \textbf{Q3}, we provide experimental results in Fig.~\ref{fig:sparsity}, where these sparsity issues are significant, especially in large-scale digraphs. 
    For feature sparsity, we assume that the feature representation of unlabeled nodes is partially missing. 
    In this case, it is necessary to obtain additional feature information from neighbors through appropriate propagation. 
    Fig.~\ref{fig:sparsity} shows that DIMPA and NSTE may suffer from limited RF due to the depth of the model, which leads to sub-optimal performance. 
    On the contrary, MagNet, MGC, and LightDiC can customize the number of propagation steps to achieve a larger RF, thus alleviating the problem of feature sparsity, which is also applicable to edge and label-sparse scenarios.
    To simulate edge sparsity, we randomly remove a fixed percentage of edges from the original digraph, providing a realistic challenge. 
    For label sparsity, we change the number of labeled samples for each class. 
    Experimental results from Fig.~\ref{fig:sparsity} show that our proposed LightDiC, as compared to baselines, is more robust to the sparsity scenarios.

\section{Conclusion}
    In this study, we propose LightDiC for real-world digraph applications.
    It follows a user-friendly decoupled paradigm and achieves impressive performance while enjoying high efficiency. 
    Remarkably, LightDiC is the only existing DiGNN that can be practically scaled to billion-level digraphs, opening avenues for future advancements.
    Through theoretical analysis, we demonstrate that LightDiC inherently optimizes smoothness, aligning with the proximal gradient descent process of the Dirichlet energy optimization function.

    \textbf{Broader Impacts and Limitations.}
    LightDiC pursues the utmost simplicity in its architecture to meet scalability challenges. 
    To improve performance, a promising direction would be to explore the node-wise RF by a well-designed attention mechanism. 
    Additionally, we use parameterized $q$ since it has been proven effective in the data-driven context. 
    However, the exploration of node-adaptive $q$ holds the potential to gain deeper insights into the inherent mechanisms underlying complex field message passing.

\appendix
\section{Outline}
The appendix is organized as follows:
\begin{description}
    
    \item[1.1] Proof of Lemma.3.1.
    \item[1.2] Proof of Lemma.3.2.
    \item[1.3] Proof of Lemma.3.3.
    \item[1.4] Proof of Lemma.3.4.
    \item[1.5] Comparison with SGC and MagNet.
    \item[1.6] Dataset Description.
    \item[1.7] Compared Baselines.
    \item[1.8] Message Aggregation Functions.
\end{description}

\subsection{Proof of Lemma.3.1}
\label{sec: Proof of quadratic form}
We represent the adjacency matrix and the diagonal degree matrix of a digraph $\mathcal{G}=(\mathcal{V}, \mathcal{E})$ with $|\mathcal{V}|=n$ nodes, $|\mathcal{E}|=m$ edges by $\mathbf{A}$, $\mathbf{D}$.
Based on this, let $\mathbf{L}_m\in\mathbb{C}^{N\times N}$ be the magnetic Laplacian of a digraph $\mathcal{G}$. 
The node feature matrix represents $\mathbf{X}\in\mathbb{C}^{N\times 1}$, the smoothness of $\mathbf{X}$ over $\mathcal{G}$ is defined as $\mathbf{X}^\dagger\mathbf{L}_m\mathbf{X}$.
Then, the total variation of the graph signal $\mathbf{X}$ is a smoothness measure, quantifying how much the signal $\mathbf{X}$ changes with respect to the graph topology encoded in magnetic Laplacian $\mathbf{L}_m$ as the following quadratic form, which is also known as Dirichlet energy 
\begin{equation}
    \begin{aligned}
    \mathbf{X}^\dagger\mathbf{L}_m\mathbf{X}=\!\!\sum_{(u,v)\in \mathcal{E}}\!\!{|\mathbf{X}_u-e^{i\theta_{uv}}\mathbf{X}_v|^2}=\!\!\sum_{(u,v)\in  \mathcal{E}}\!\!{|e^{-i\theta_{uv}}\mathbf{X}_u-\mathbf{X}_v|^2}.
    \end{aligned}
\end{equation}

First, by expanding $\mathbf{L}_m\mathbf{X}$, we have \begin{equation}
    \left(\mathbf{L}_m\mathbf{X}\right)_k=\!\!\!\!\sum_{w\in \mathcal{N}^{in}(k)\cup \mathcal{N}^{out}(k)}\!\!\!\!{\left(\mathbf{X}_k-e^{i\theta_{kw}}\mathbf{X}_w\right)}.
\end{equation} 
Based on this, $\mathbf{X}^\dagger\mathbf{L}_m\mathbf{X}$ can be formulated as:
\begin{equation}
\begin{split}
    \mathbf{X}^\dagger\mathbf{L}_m\mathbf{X}
    &=\sum_{k\in \mathcal{V}}\!\!\sum_{w\in \mathcal{N}^{in}(k)\cup \mathcal{N}^{out}(k)}\!\!{\left(\mathbf{X}_k-e^{i\theta_{kw}}\mathbf{X}_w\right)}\\
    &=\!\!\!\sum_{(u,v)\in \mathcal{E}}\!\!\!{\mathbf{X}_u^\dagger\mathbf{X}_u-e^{i\theta_{uv}}\mathbf{X}_u^\dagger\mathbf{X}_v+\mathbf{X}_v^\dagger\mathbf{X}_v-e^{i\theta_{vu}}\mathbf{X}_v^\dagger\mathbf{X}_u}\\
    &=\!\!\!\sum_{(u,v)\in \mathcal{E}}\!\!\!{\mathbf{X}_u^\dagger\mathbf{X}_u+\mathbf{X}_v^\dagger\mathbf{X}_v-\left(e^{i\theta_{uv}}\mathbf{X}_u^\dagger\mathbf{X}_v+e^{i\theta_{vu}}\mathbf{X}_v^\dagger\mathbf{X}_u\right)}\\
    &=\!\!\!\sum_{(u,v)\in \mathcal{E}}\!\!\!|\mathbf{X}_u|^2+|\mathbf{X}_v|^2-\\
    &\;\;\;\;\;\;\;|\mathbf{X}_u||\mathbf{X}_v|\left(e^{i\left(-\theta_{\mathbf{X}_u}+\theta_{\mathbf{X}_v}+\theta_{uv}\right)}+e^{i\left(-\theta_{\mathbf{X}_v}+\theta_{\mathbf{X}_u}+\theta_{vu}\right)}\right)\\
    &=\!\!\!\!\!\sum_{(u,v)\in \mathcal{E\!\!}}\!\!\!{|\mathbf{X}_u|^2\!+\!|\mathbf{X}_v|^2\!-\!2|\mathbf{X}_u||\mathbf{X}_v|\cos\left(\theta_{\mathbf{X}_u}-\theta_{\mathbf{X}_v}-\theta_{uv}\right)},
\end{split}
\end{equation}
the last equality holds because $\theta_{uv}=-\theta_{vu}$.
Since 
\begin{equation}
\begin{split}
    |\mathbf{X}_u-e^{i\theta_{uv}}\mathbf{X}_v|^2&=||\mathbf{X}_u|e^{i\theta_{\mathbf{X}_u}}-e^{i\theta_{uv}}|\mathbf{X}_v|e^{i\theta_{\mathbf{X}_v}}|^2\\
    &=||\mathbf{X}_u|e^{i\theta_{\mathbf{X}_u}}-|\mathbf{X}_v|e^{i(\theta_{\mathbf{X}_v}+\theta_{uv})}|^2\\
    &=||\mathbf{X}_u|\cos\theta_{\mathbf{X}_u}+i|\mathbf{X}_u|\sin\theta_{\mathbf{X}_u}-\\
    &\;\;\;\;\left(|\mathbf{X}_v|\cos\left(\theta_{\mathbf{X}_v}+\theta_{uv}\right)+i|\mathbf{X}_v|\sin\left(\theta_{\mathbf{X}_v}+\theta_{uv}\right)\right)|^2\\
    &=||\mathbf{X}_u|\cos\theta_{\mathbf{X}_u}-|\mathbf{X}_v|\cos(\theta_{\mathbf{X}_v}+\theta_{uv})+\\
    &\;\;\;\;\left(|\mathbf{X}_u|\sin\theta_{\mathbf{X}_u}-|\mathbf{X}_v|\sin\left(\theta_{\mathbf{X}_v}+\theta_{uv}\right)\right)|^2\\
    &=\left(|\mathbf{X}_u|\cos\theta_{\mathbf{X}_u}-|\mathbf{X}_v|\cos(\theta_{\mathbf{X}_v}+\theta_{uv})\right)^2+\\
    &\;\;\;\;\left(|\mathbf{X}_u|\sin\theta_{\mathbf{X}_u}-|\mathbf{X}_v|\sin\left(\theta_{\mathbf{X}_v}+\theta_{uv}\right)\right)^2\\
    &=|\mathbf{X}_u|^2+|\mathbf{X}_v|^2-\\
    &\;\;\;\; 2|\mathbf{X}_u||\mathbf{X}_v|\cos\left(\theta_{\mathbf{X}_u}-\theta_{\mathbf{X}_v}-\theta_{uv}\right)\\
    &=|e^{-i\theta_{uv}}\mathbf{X}_u-\mathbf{X}_v|^2,
\end{split}
\end{equation} 
the lemma is proved.

\subsection{Proof of Lemma.3.2}
\label{sec: Proof of low-pass}
In LightDiC, we predefine the magnetic graph operator
\begin{equation}
    \begin{aligned}
\text{MGO}:=\hat{\mathbf{A}}_m=\left(\widetilde{\mathbf{D}}_m^{-1/2}\widetilde{\mathbf{A}}_m\widetilde{\mathbf{D}}_m^{-1/2} \odot \exp \left(i \Theta^{(q)}\right)\right).
    \end{aligned}
\end{equation}
MGO is a digraph operator defined in terms of the global topology based on the complex domain,  modeling the directional information through the imaginary part.
Based on this, we can define complex domain-based message-passing mechanisms for digraphs.
Now we further analyze the digraph operator used in our proposed LightDiC: $\left(\widetilde{\mathbf{D}}_m^{-1/2}\widetilde{\mathbf{A}}_m\widetilde{\mathbf{D}}_m^{-1/2} \odot \exp \left(i \Theta^{(q)}\right)\right)$.

As an analogy of spectral graph convolution on undirected graphs, the directed spectral graph convolution is also based on the convolution theorem. 
The Fourier transform for a signal $\mathbf{X}:\mathbb{V}\rightarrow \mathbb{C}$ is defined as $\widehat{\mathbf{X}}=\mathbf{U}^\dagger\mathbf{X}$, so that $\widehat{\mathbf{X}}(k)=\left<\mathbf{X},\mathbf{u}_k\right>$, where $\mathbb{V}$ and $\mathbb{C}$ represents vertex domain an complex field-based Fourier domain respectively. 
Note that the elements of $\mathbf{u}_1,\cdots,\mathbf{u}_n$ are complex numbers, obtained by eigendecomposition of the magnetic Laplacian $\mathbf{L}_m$, we regard $\mathbf{u}_1,\cdots,\mathbf{u}_n$ as the Fourier basis for digraphs. 
Since $\mathbf{U}$ is unitary, we have the Fourier inverse formula:
\begin{equation}\label{equation:inverse-formula}
\mathbf{X}=\mathbf{U}\widehat{\mathbf{X}}=\sum_{k=1}^N{\widehat{\mathbf{X}}(k)\mathbf{u}_k}.
\end{equation}
Here, convolution corresponds to point-wise multiplication on the Fourier basis. 
Thus, the convolution of $\mathbf{X}$ with a filter $\mathbf{g}$ in the Fourier domain can be defined as $\widetilde{\mathbf{g}\star\mathbf{X}}(k)=\widetilde{\mathbf{g}}(k)\widetilde{\mathbf{X}}(k)$. 
According to Eq.(\ref{equation:inverse-formula}), $\mathbf{g}\star\mathbf{X}=\mathbf{U}\text{Diag}(\widehat{\mathbf{g}})\widehat{\mathbf{X}}=(\mathbf{U}\text{Diag}(\widehat{\mathbf{g}})\mathbf{U}^\dagger)\mathbf{X}$, the convolution matrix can be written as:
\begin{equation}
    \mathbf{G}=\mathbf{U}\Sigma\mathbf{U}^\dagger,
\end{equation}
For a diagonal matrix $\Sigma$, different filter refers to different choices of $\Sigma$. In practice, in order to avoid explicit eigen-decomposition, $\Sigma$ is often set as polynomials of $\Lambda$. Suppose $\widetilde{\Lambda}$ is a normalized eigenvalue matrix, which is defined as $\widetilde{\Lambda}=\frac{2}{\lambda_{\text{max}}}\Lambda-\mathbf{I}$, we can write
\begin{equation}
    \mathbf{\Sigma}=\sum_{k=0}^K{\mathbf{w}_k\operatorname{T}_k(\widetilde{\Lambda})}.
\end{equation}
$\operatorname{T}_k$ can be set as various orthogonal polynomial bases. 
When choosing $\operatorname{T_k}$ as the $k$-order Chebyshev polynomial which is defined by $\operatorname{T}_0(x)=1$, $\operatorname{T}_1(x)=x$, and $\operatorname{T}_k(x)=2x\operatorname{T}_{k-1}(x)+\operatorname{T}_{k-2}(x)$ for $k\geq2$. Since $(\mathbf{U}\widetilde{\Lambda}\mathbf{U}^\dagger)^k=\mathbf{U}\widetilde{\Lambda}^k\mathbf{U}^\dagger$, we have:
\begin{equation}
\mathbf{GX}=\mathbf{U}\sum_{k=0}^K{\mathbf{w}_k\operatorname{T}_k(\widetilde{\Lambda})\mathbf{U}^\dagger\mathbf{X}}=\sum_{k=0}^K\mathbf{w}_k\operatorname{T}_k(\widetilde{{\mathbf{L}}}_m)\mathbf{X}.
\end{equation}
Here, $\widetilde{{\mathbf{L}}}_m$ is defined as $\widetilde{{\mathbf{L}}}_m=\frac{2}{\lambda_{\text{max}}}{\mathbf{L}_m}-\mathbf{I}$. 
We set $K=1$, assume that $\mathbf{L}_m=\mathbf{L}_m^{(q)}$, $\lambda_{\text{max}}\approx2$ and $\mathbf{w}=\mathbf{w}_0=-\mathbf{w}_1$, we can obtain that:
\begin{equation}
\mathbf{GX}=\mathbf{W}\left(\mathbf{I}+\mathbf{D}_m^{-1/2}\mathbf{A}_m\mathbf{D}_m^{-1/2} )\odot \exp \left(i \Theta^{(q)}\right)\right)\mathbf{X}.
\end{equation}
By applying further renormalization tricks to avoid instabilities arising from vanishing/exploding gradients, it yields
\begin{equation}
\mathbf{GX}=\mathbf{W}\left(\widetilde{\mathbf{D}}_m^{-1/2}\widetilde{\mathbf{A}}_m\widetilde{\mathbf{D}}_m^{-1/2} \odot \exp \left(i \Theta^{(q)}\right)\right)\mathbf{X},
\end{equation}
where $\widetilde{\mathbf{A}}_m=\mathbf{A}_m+\mathbf{I}$ and $\widetilde{\mathbf{D}}_m(i,i)=\sum_j{\widetilde{\mathbf{A}}_m(i,j)}$.

According to the analysis, we can find that the complex domain feature propagation in our proposed LightDiC is a complex spectral convolution with a low-pass filter.

Since $\hat{\mathbf{A}}_m^{K}\mathbf{X}=\left(\widetilde{\mathbf{D}}_m^{-1/2}\widetilde{\mathbf{A}}_m\widetilde{\mathbf{D}}_m^{-1/2} \odot \exp \left(i \Theta^{(q)}\right)\right)\mathbf{X}$, we have $\hat{\mathbf{A}}_m=\widetilde{\mathbf{D}}_m^{-1/2}\widetilde{\mathbf{A}}_m\widetilde{\mathbf{D}}_m^{-1/2} \odot \exp \left(i \Theta^{(q)}\right)\approx\mathbf{I}-\mathbf{L}_m$. 
Then, $\hat{\mathbf{A}}_m^{K}=(\mathbf{I}-\mathbf{L}_m)^K=\mathbf{U}(\mathbf{I}-{\Sigma})^K\mathbf{U}^\dagger$, applying this operation on the digraph signal $\mathbf{X}$ ($\hat{\mathbf{A}}_m\mathbf{X}=\mathbf{U}\left(\mathbf{I}-\Sigma\right)^K\mathbf{U}^\dagger$) is identical to:
(i) convert $\mathbf{X}$ to the Fourier domain ($\mathbf{U}^\dagger\mathbf{X}$); 
(ii) applying a low-pass filter $g(\lambda_i)=(1-\lambda_i)^K$ for $i=1,\cdots,n$ on the signal on the Fourier domain ($(\mathbf{I}-\Sigma)^K\mathbf{U}^\dagger\mathbf{X}$); 
(iii) applying inverse Fourier transform on the signal ($\mathbf{U}(\mathbf{I}-\Sigma)^K\mathbf{U}^\dagger\mathbf{X}$). 

The function $(\mathbf{I}-\mathbf{X})^K$ is a monotonically decreasing function. 
Since each signal $\mathbf{X}\in\mathbb{C}^n$ can be written as $\mathbf{X}=\sum_{i=1}^n{\omega_i \mathbf{u}_i}$.
The operation will keep the information of eigenvector components corresponding to the lower eigenvalues and ignore the components corresponding to the higher eigenvalues.

\subsection{Proof of Lemma.3.3}
\label{sec: Proof of low-eigen}
Since in real-world graphs, the feature varies smoothly over the graph, an intuition of our feature pre-processing is to smooth signals. 
To evaluate the smoothness of $\hat{\mathbf{A}}_m^{K}\mathbf{x}$, we first analyze the smoothness of the eigenvectors of $\hat{\mathbf{A}}_m$.

\begin{figure*}[t]   
	\centering
    \setlength{\abovecaptionskip}{0.4cm}
	\includegraphics[width=\linewidth,scale=1.00]{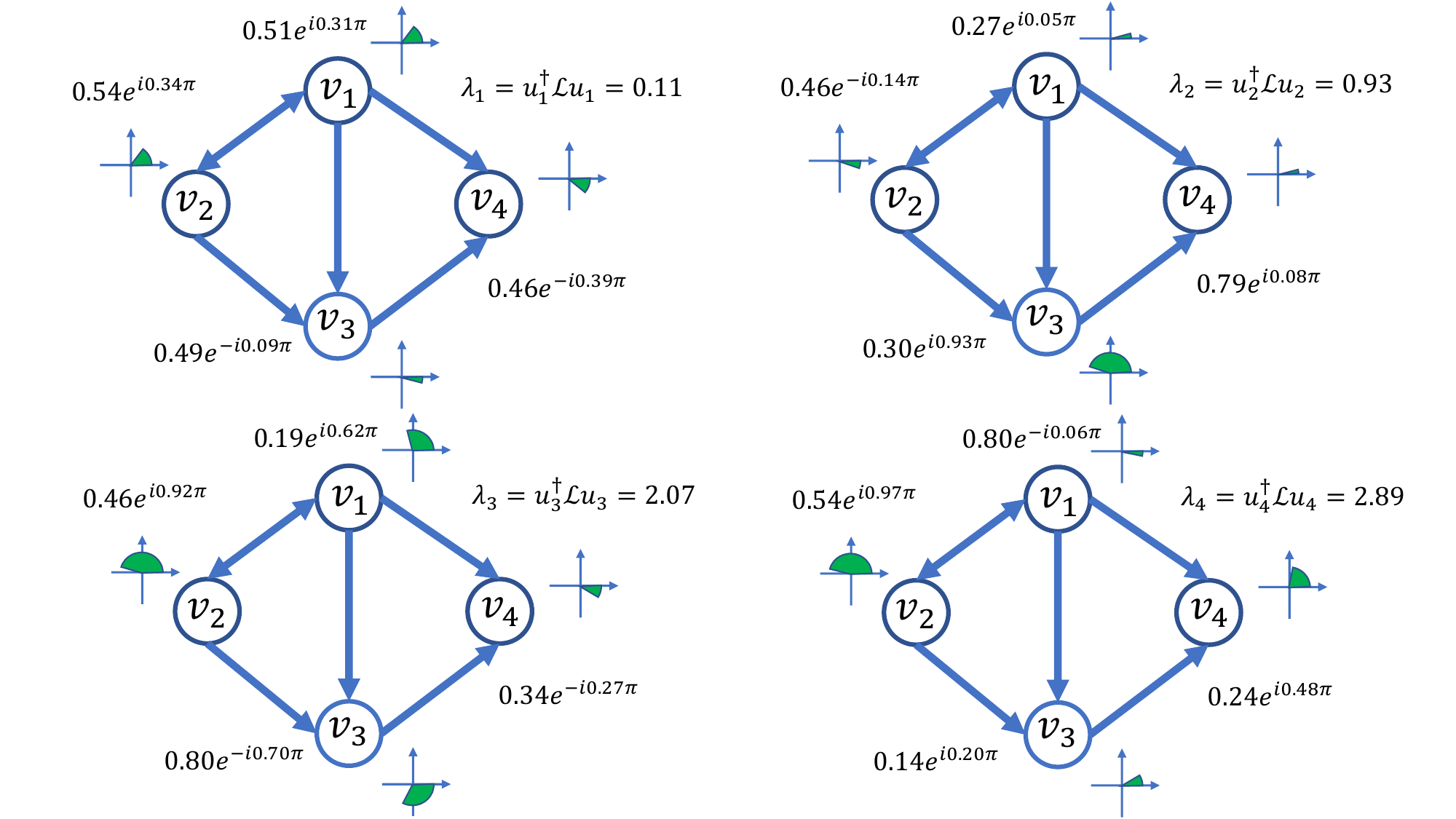}
	\caption{
 Illustration of all the eigenvectors of a magnetic Laplacian $\mathbf{L}^{(q)}_m=\mathbf{D}_m-\mathbf{A}_m \odot \exp \left(i \Theta^{(q)}\right)$ with $q=0.25$ of an example graph $\mathcal{G}$. 
 The eigenvectors corresponding to smaller eigenvalues vary more smoothly on graphs. 
 When evaluating the smoothness of a signal $\mathbf{X}$ as $\mathbf{X}^\dagger\mathcal{L}\mathbf{X}$, the smoothness of $\mathbf{u}_1=[0.51e^{i0.31\pi},0.54e^{i0.34\pi},0.49e^{-i0.09\pi},0.46e^{-i0.39\pi}]^{\operatorname{T}}$ is $\lambda_1=0.11$, followed by the smoothness of $\mathbf{u}_2$ ($0.93$), $\mathbf{u}_3$ ($2.07$) and $\mathbf{u}_4$ ($2.89$).}
	\label{fig:low_eigen}
\end{figure*}

According to Courant-Fischer theorem theorem~\cite{cf_theorem1}, we have
\begin{lemma}
    Let $\mathbf{L}_m\in\mathbb{C}^{n\times n}$ be a Hermitian matrix with eigenvalues $\lambda_1\leq\cdots\leq\lambda_n$, then we have
    \begin{equation}
        \lambda_k=\max_{S\in\mathbb{C}^n,\ dim(S)=k}\min_{\mathbf{X}\in S}\frac{ \mathbf{X}^\dagger\mathbf{L}_m\mathbf{X}}{\mathbf{X}^\dagger\mathbf{X}}
    \end{equation}
\end{lemma}
The smoothness metric $\mathbf{X}^\dagger\mathbf{L}_m\mathbf{X}$ is minimized by the eigenvector $\mathbf{u}_1$ corresponding to the smallest eigenvalue. 
Furthermore, we can expect that smaller eigenvalues have corresponding eigenvectors with smaller smoothness. 
We provide an example digraph $\mathcal{G}$ to prove our point, the four eigenvectors of the magnetic Laplacian $\mathbf{L}^{(q)}_m:=\mathbf{D}_m-\mathbf{A}_m^{(q)}=\mathbf{D}_m-\mathbf{A}_m \odot \exp \left(i \Theta^{(q)}\right)$ with $q=0.25$ of $\mathcal{G}$ is illustrated in Fig.~\ref{fig:low_eigen}. 
It shows that the magnetic Laplaican of $\mathcal{G}$ has 4 eigenvectors $\mathbf{u}_1$, $\mathbf{u}_2$, $\mathbf{u}_3$, $\mathbf{u}_4\in\mathbb{C}^n$.
The elements of these four eigenvectors are all complex numbers, we represent a complex number $x$ by the magnitude $|x|$ and the angle $\theta=\text{argtan}\left(\text{Imag}(x)/\text{Real}(x)\right)$. 
Each eigenvector $\mathbf{u}_k$ can be regarded as a complex graph signal on graphs, with $\mathbf{u}_k(i)$ being the complex signal on node $i\in \mathbb{V}$. 
We depict the value of the four eigenvectors in Figure.\ref{fig:low_eigen}, we also plot the angle of each eigenvector element. 
Then, the smoothness of each eigenvector can be computed as $\mathbf{u}_k^\dagger\mathbf{L}_m\mathbf{u}_k$ according to our definition. 
Since $\mathbf{L}_m\mathbf{u}_k=\lambda_k\mathbf{u}_k$, $\lambda_k=\mathbf{u}_k^\dagger\mathbf{L}_m\mathbf{u}_k$, the smoothness value exactly equals the eigenvalue. 
$\mathbf{u}_1$ corresponding to the smallest eigenvalue has the smallest smoothness value, followed by $\mathbf{u}_2$, $\mathbf{u}_3$ and $\mathbf{u}_4$. 
It can be seen that the eigenvectors corresponding to smaller eigenvalues vary more smoothly on graphs.
As a result, the low eigen-components of the node feature also vary very smoothly over the graph. 
Thus, the complex domain propagation keeps the low-frequency components of the feature and ignores the high-frequency components. 
The resulting signal is smooth while keeping the information of the node feature.

\subsection{Proof of Lemma.3.4}
\label{sec: Proof of pgd}
Given digraph signal $\mathbf{y}\in\mathbb{C}^{N\times 1}$ with noises ${\epsilon}$, the optimization function is defined as $\min_{\mathbf{X}}f(\mathbf{X})=\min_{\mathbf{X}\in\mathbb{C}}\Vert\mathbf{X}-\mathbf{y}\Vert_2+\mathbf{X}^\dagger\mathbf{L}_m\mathbf{X}.$

In order to solve a convex optimization problem $\min_{\mathbf{X}}f(\mathbf{X})$, the proximal gradient descent method updates the current solution $\mathbf{X}^{(t)}$ by $\mathbf{X}^{(t+1)}\leftarrow\mathbf{X}^{(t)}+\lambda\nabla f(\mathbf{X}^{(t)})$ where $\lambda$ is the step size and $\nabla f(\mathbf{X})$ is the gradient of $f(\mathbf{X})$ w.r.t. $\mathbf{X}$. Since $\mathbf{Z}(\mathbf{X})=\min_{\mathbf{X}\in\mathbb{C}}\Vert\mathbf{X}-\mathbf{y}\Vert_2+\mathbf{X}^\dagger\mathbf{L}_m\mathbf{X}$, we have:
\begin{equation}
    \begin{split}
        \frac{\partial \mathbf{Z}(\mathbf{X})}{\partial \mathbf{X}}=2\mathbf{L}_m\mathbf{X}+2\mathbf{X}-2\mathbf{y}.
    \end{split}
\end{equation}
Then, start from a initial solution $\mathbf{X}^{(0)}=\mathbf{X}$, the $k+1$ step gradient descent can be computed as:
\begin{equation}
    \begin{split}
    \mathbf{X}^{(k+1)}=\mathbf{X}^{(k)}-\alpha[(\mathbf{L}_m+\mathbf{I})\mathbf{X}^{(k)}-\mathbf{y}],
    \end{split}
\end{equation}
where $\lambda=\frac{\alpha}{2}$ is the step size. 
In practice, the convergence speed of such an update strategy is often slow, pre-conditioning is a well-known technique in numerical optimization to accelerate the convergence. 
By applying pre-conditioning and re-arranging the terms, we have:
\begin{equation}
    \begin{split}
        \mathbf{X}^{(k+1)}=(1-\alpha)\mathbf{X}^{(k)}+\alpha\widetilde{\mathbf{D}}_m^{-1}[\mathbf{A}_m\mathbf{X}^{(k)}+\mathbf{y}],
    \end{split}
\end{equation}
where $\widetilde{\mathbf{D}}_m^{-1}=(\mathbf{D}_m+\mathbf{I})^{-1}$. 
Therefore, start from an initial solution $\mathbf{X}^{(0)}=\mathbf{D}_m^{1/2}\mathbf{X}$, after the first gradient step, $\mathbf{X}^{(1)}$ is equivalent to the one layer model $\widetilde{\mathbf{D}}_m^{-1/2}\widetilde{\mathbf{A}}_m\widetilde{\mathbf{D}}_m^{-1/2}\mathbf{X}$ up to a simple re-parameterization by $\widetilde{\mathbf{D}}_m^{-1/2}$. 
Thus, the $K$-step propagated features $\hat{\mathbf{A}}_m^{K}\mathbf{X}=\left(\widetilde{\mathbf{D}}_m^{-1/2}\widetilde{\mathbf{A}}_m\widetilde{\mathbf{D}}_m^{-1/2} \odot \exp \left(i \Theta^{(q)}\right)\right)\mathbf{X}$ is equivalent to applying such a proximal gradient update for $K$ times.

Up to this point, we have presented the theoretical extension of spectral GNNs to digraphs using the magnetic Laplacian. 
This integration offers a unified theoretical framework encompassing smoothness, Dirichlet energy, and spectral analysis.

\subsection{Comparison with SGC and MagNet.}
\textbf{Comparison with SGC~\cite{wu2019sgc}.}
    LightDiC differs from the related work SGC in:
    (i) Application-guided Generalization: SGC is limited in handling undirected graphs, and our experiments have revealed its significant performance degradation in large-scale digraphs. 
    In contrast, LightDiC focuses more on digraphs with complex topologies and aims to provide a universal solution for both directed and undirected graphs through MGO.
    (ii) Paradigm-guided Scalability: SGC is recognized for introducing the decoupling design paradigm to address scalability challenges posed by large-scale graph data.
    Inspired by this, subsequent undirected methods like SIGN~\cite{frasca2020sign}, S$^2$GC~\cite{zhu2021ssgc}, GBP~\cite{chen2020gbp}, GAMLP~\cite{gamlp}, and GRAND+~\cite{feng2022grand+} have further improved performance through well-designed learnable mechanisms. 
    However, LightDiC is the first to successfully apply this decoupling paradigm to large-scale digraphs, offering a novel attempt to address real-world data science challenges.
    (iii) Technical-guided Innovation: SGC relies on symmetrically normalized adjacency matrices for smoothing signals on undirected graphs and utilizes the propagated features from the final step for downstream task training.
    Comparatively, LightDiC employs the magnetic Laplacian to smooth signals on digraphs in the complex number domain. 
    We theoretically demonstrate that both the real and imaginary parts contain rich topological insights. 
    Consequently, LightDiC proposes a weight-free message aggregation function that combines information from these two different encoding perspectives.

\textbf{Comparison with MagNet~\cite{zhang2021magnet}.}
    LightDiC differs from MagNet in:
    (i) Scalability: While MagNet achieves efficient convolutions on digraphs using the magnetic Laplacian, it inherits unnecessary computational redundancy from deep learning (complex computations and more learnable parameters). 
    Additionally, MagNet's model depth and applicability are limited to shallow design and toy-sized datasets due to the recursive computations in the complex number domain, lacking scalability. 
    In contrast, LightDiC minimizes precomputation overhead through a decoupling design paradigm, making it solely dependent on efficient matrix multiplication based on sparse matrices. 
    LightDiC also compresses complex learning processes into simple linear mappings to maximize training and inference efficiency.
    (ii) Encoding Strategy: Unlike MagNet, which employs recursive GCN as the encoding strategy, LightDiC introduces a weight-free message aggregation function based on appropriate theoretical extensions on digraphs. 
    This function encodes multi-scale deep structural information, enhancing predictive performance.
    (iii) Supplemental Interpretability: MagNet theoretically proves that the symmetrically conjugated magnetic Laplacian shares properties similar to symmetrically normalized undirected graph Laplacians. 
    It also provides empirical analysis of its eigenvalues and eigenvectors. 
    Building upon this, LightDiC extends theoretical spectral analysis from undirected to digraphs from the perspective of graph signal denoising.
    It uses this perspective to guide the design of suitable encoding mechanisms. 
\subsection{Dataset Description}
\label{sec: Dataset Description}
    The description of all digraph benchmark datasets is listed below:
    
    \textbf{CoraML}, \textbf{CiteSeer}~\cite{bojchevski2018coraml_citeseer}, and \textbf{ogbn-papers100M}~\cite{hu2020ogb} are three citation network datasets. 
    In these three networks, papers from different topics are considered nodes, and the edges are citations among the papers. 
    The node attributes are binary word vectors, and class labels are the topics the papers belong to.
    
    \textbf{WikiCS}~\cite{mernyei2020wikics} is a Wikipedia-based dataset for benchmarking GNNs. 
    The dataset consists of nodes corresponding to computer science articles, with edges based on hyperlinks and 10 classes representing different branches of the field.
    The node features are derived from the text of the corresponding articles. 
    They were calculated as the average of pre-trained GloVe word embeddings~\cite{pennington-etal-2014-glove}, resulting in 300-dimensional node features.
    
    \textbf{Slashdot}~\cite{ordozgoiti2020slashdot} is from a technology-related news website with user communities. 
    The website introduced Slashdot Zoo features that allow users to tag each other as friends or foes. 
    The dataset is a common signed social network with friends and enemies labels.
    In our experiments, we only consider friendships.
    
    \textbf{Epinions}~\cite{massa2005epinions} is a who-trust-whom online social network. 
    Members of the site can indicate their trust or distrust of the reviews of others. 
    The network reflects people's opinions of others.
    In our experiments, we only consider the "trust" relationships.
    
    \textbf{WikiTalk}~\cite{leskovec2010wikitalk} contains all users and discussions from the inception of Wikipedia until Jan. 2008. 
    Nodes in the network represent Wikipedia users and a directed edge from node $v_i$ to node $v_j$ denotes that user $i$ edited at least once a talk page of user $j$. 
    
    \subsection{Compared Baselines}
    \label{sec: Compared Baselines}
    The main characteristics of all baselines are listed below
    
    \textbf{DGCN}~\cite{tong2020dgcn}: DGCN proposes the first and second-order proximity of neighbors to design a new message-passing mechanism, which in turn learns aggregators based on incoming and outgoing edges using two sets of independent learnable parameters.
    
    \textbf{DIMPA}~\cite{he2022dimpa}: DIMPA represents source and target nodes separately. However, DIMPA aggregates the neighborhood information within $K$ hops in each layer to further increase the receptive field (RF), and it performs a weighted average of the multi-hop neighborhood information to capture the local network information.
    
    \textbf{NSTE}~\cite{kollias2022nste}: NSTE is inspired by the 1-WL graph isomorphism test, which uses two sets of trainable weights to encode source and target nodes separately. Then, the information aggregation weights are tuned based on the parameterized feature propagation process to generate node representations.
    
    \textbf{DiGCN}~\cite{tong2020digcn}: DiGCN notices the inherent connections between graph Laplacian and stationary distributions of PageRank, it theoretically extends personalized PageRank to construct real symmetric Digraph Laplacian. Meanwhile, DiGCN uses first-order and second-order neighbor proximity to further increase RF.
    
    \textbf{DiGCN-Appr}~\cite{tong2020digcn}: DiGCN with fast personalized PageRank approximation, without inception blocks, which can be viewed as the generalization of GCN based on the digraph Laplacian.
    
    \textbf{DiGCN-IB}~\cite{tong2020digcn}: DiGCN with inception blocks, without fast personalized PageRank approximation, which can be viewed as an optimized message-passing mechanism.

    \textbf{MagNet}~\cite{zhang2021magnet}: MagNet utilizes complex numbers to model directed information, it proposes a spectral GNN for digraphs based on a complex Hermitian matrix known as the magnetic Laplacian. Meanwhile, MagNet uses additional trainable parameters to combine the real and imaginary filter signals separately to achieve better prediction performance.
    
    \textbf{MGC}~\cite{zhang2021mgc}: MGC introduces the magnetic Laplacian, a discrete operator with the magnetic field, which preserves edge directionality by encoding it into a complex phase with an electric charge parameter. 
    By adopting a truncated variant of PageRank named Linear-Rank, it design and build a low-pass filter for homogeneous graphs and a high-pass filter for heterogeneous graphs.

   \textbf{GCN}~\cite{kipf2016gcn}: GCN is guided by a localized first-order approximation of spectral graph convolutions. This model's scalability is directly proportional to the number of graph edges, and it learns intermediate representations in hidden layers that capture both the local graph arrangement and node-specific features.
   
    \textbf{GraphSAGE}~\cite{hamilton2017graphsage}: GraphSAGE is a scalable and flexible approach for node embedding generation in large graphs, which leverages a sampling-based neighborhood aggregation scheme and a multi-layer perceptron to generate embeddings that capture both local and global structural information.
    
    \textbf{UniMP}~\cite{shi2020unimp}: UniMP adopts the Graph Transformer model as the base model and then combines label embedding to feed node features and part of node labels into the model in parallel, which enables the propagation of node features and labels.
    
    \textbf{SGC}~\cite{wu2019sgc}: SGC is a simple and efficient graph convolutional network that applies a fixed graph filter to the input features. 
    It simplifies GCN by removing nonlinearities and collapsing weight matrices between consecutive layers, which achieves competitive performance on various graph-related tasks with significantly reduced computational cost compared to other GCN models.
    
    \textbf{SIGN}~\cite{frasca2020sign}: SIGN is a scalable GNN that uses an inception module to learn hierarchical representations of nodes in large-scale graphs. 
    It achieves state-of-the-art performance on various graph classification tasks and is scalable to graphs with millions of nodes.
    
    \textbf{GBP}~\cite{chen2020gbp}: GBP lies in its bidirectional propagation mechanism, a process that computes a versatile Generalized PageRank matrix to capture a wide spectrum of graph convolutions, thereby expanding the horizon of expressive power within graph-based operations.
    
    \textbf{S$^2$GC}~\cite{zhu2021ssgc}: S$^2$GC uses a modified Markov Diffusion Kernel to generalize GCN, and it can be used as a trade-off of low and high-pass filter which captures the global and local contexts of nodes.
    
    \textbf{GAMLP}~\cite{gamlp}: GAMLP introduces a duo of novel attention mechanisms: recursive attention and JK attention, revolutionizing the way representations are learned over RF of varying sizes in a dynamic, node-specific fashion. 

\begin{table}[]
\setlength{\abovecaptionskip}{0.2cm}
\setlength{\belowcaptionskip}{-0.2cm}
\caption{Performance on different message aggregation functions. 
}
\label{tab: message aggregation}
\resizebox{\linewidth}{29mm}{
\setlength{\tabcolsep}{1mm}{
\begin{tabular}{cc|cccc}
\hline
Datasets                                                                   & Tasks     & Last              & Mean              & Sum               & Concat            \\ \midrule[0.3pt]
\multirow{4}{*}{CoraML}                                                    & Node-C    & \textbf{84.2±0.7} & 83.6±0.5          & 84.0±0.6          & 83.9±0.5          \\
                                                                           & Existence & 78.9±0.2          & 79.6±0.1          & \textbf{80.3±0.2} & 80.1±0.2          \\
                                                                           & Direction & 90.3±0.2          & 90.5±0.3          & \textbf{90.7±0.3} & 90.6±0.2          \\
                                                                           & Link-C    & 74.1±0.2          & 74.3±0.1          & 74.6±0.1          & \textbf{74.8±0.3} \\ \midrule[0.3pt]
\multirow{4}{*}{WikiCS}                                                    & Node-C    & \textbf{79.8±0.4} & 79.2±0.2          & 79.6±0.3          & 79.5±0.2          \\
                                                                           & Existence & 87.8±0.0          & \textbf{85.6±0.1} & 85.5±0.0          & 85.3±0.1          \\
                                                                           & Direction & 88.4±0.1          & 88.7±0.1          & \textbf{89.2±0.1} & 89.0±0.1          \\
                                                                           & Link-C    & 79.5±0.2          & 80.2±0.1          & 80.1±0.1          & \textbf{80.4±0.2} \\ \midrule[0.3pt]
\multirow{4}{*}{\begin{tabular}[c]{@{}c@{}}ogbn\\ papers100M\end{tabular}} & Node-C    & 65.2±0.2          & 64.6±0.1          & 64.8±0.1          & \textbf{65.4±0.2} \\
                                                                           & Existence & 90.7±0.0          & 91.3±0.0          & \textbf{91.6±0.1} & 91.2±0.1          \\
                                                                           & Direction & 93.0±0.0          & 93.2±0.1          & \textbf{93.8±0.1} & 93.4±0.0          \\
                                                                           & Link-C    & 89.4±0.1          & 89.8±0.1          & 90.0±0.1          & \textbf{90.3±0.1} \\ \midrule[0.3pt]
\end{tabular}
}}
\vspace{-0.3cm}
\end{table}

\vspace{-0.2cm}
\subsection{Message Aggregation Functions}
\label{sec: Message Aggregation Functions}
    In this section, we extend our analysis of LightDiC by conducting experiments with different message aggregation functions. 
    This expansion serves to further validate the viewpoints presented in Sec~3.1 of our main text:
    (1) Utilizing a weight-based method inspired by SIGN~\cite{frasca2020sign} to merge real and imaginary features could negatively impact predictive performance. 
    This is attributed to the distinct physical meanings of the real and imaginary components. 
    Employing a single learnable weight to combine them indiscriminately is ill-advised due to the lack of consideration for their individual characteristics.
    Conversely, a weight-free approach, while simple, can have a positive impact on prediction by intuitively encoding deep structural information.
    (2) Building upon the aforementioned point, in our proposed LightDiC, we opt for a weight-free message aggregation function to maintain simplicity, avoiding additional computational overhead caused by separately considering real and imaginary components, as well as the intricacies of a model architecture that combines both. 
    This choice facilitates the handling of propagated features obtained from performing $K$-step propagation based on the MGO.
    For the weight-free message aggregation functions based on the $K$-step graph propagation, we consider the following:
    (i) $\operatorname{Last}(\cdot)$ directly selects the propagated feature matrix obtained at the $K$-th step as the output.
    (ii) $\operatorname{Mean}(\cdot)$ computes the element-wise average of propagated features;
    (iii) $\operatorname{Sum}(\cdot)$ calculates the element-wise sum of propagated features;
    (iiii) $\operatorname{Concat}(\cdot)$ concatenates $K$ propagated feature matrices;

     Building upon the experimental findings in Table~\ref{tab: message aggregation}, the following conclusions can be drawn:
     (1) For node-level tasks, the choice of optimal aggregation functions is contingent upon the dimensional of node features provided by the original dataset. 
     For instance, in the case of feature-rich datasets such as CoraML and WikiCS, node prediction performance relies on smoothed features. 
     Consequently, selecting $\operatorname{Last}(\cdot)$ yields the best performance by attaining maximal smoothing signals. 
     Conversely, for ogbn-papers100M with only 128 dimensions for node features, $\operatorname{Concat}(\cdot)$ is necessary to enhance predictive performance through concatenating multi-scale node smoothing representations.
    (2)For link-level tasks, the enhancement of predictive performance necessitates capturing the intricate topological structure of the digraph. 
    Thus, operators such as $\operatorname{Mean}(\cdot)$, $\operatorname{Sum}(\cdot)$, and $\operatorname{Concat}(\cdot)$, which encode multi-dimensional deep structural information, tend to outperform $\operatorname{Last}(\cdot)$ significantly. 
    Additionally, because $\operatorname{Mean}(\cdot)$ mitigates difference between propagated features, which results in less distinguishable final node representations, its effectiveness falls short compared to $\operatorname{Sum}(\cdot)$. 
    It is worth noting that as Link-C demands more fine-grained classification requirements, the enriched node representations generated by $\operatorname{Concat}(\cdot)$ offer substantial benefits for its performance.

\newpage
\newpage
\balance{
\bibliographystyle{ACM-Reference-Format}
\bibliography{sample}
}


\end{document}